\documentclass[12pt]{article}
\usepackage{amsmath}
\usepackage{amsthm}
\usepackage{amssymb}
\usepackage{color}
\usepackage{graphicx,psfrag,epsf}
\usepackage{enumerate}
\usepackage{natbib}
\usepackage{subfigure}
\usepackage{authblk}
\usepackage{parskip}
\usepackage{url} 

\newcommand{\blind}{0}

\newcommand{\M}[1]{\boldsymbol{#1}}  
\newcommand{\V}[1]{\boldsymbol{#1}}  
\newcommand{\EX}[0]{\mathbb{E}}

\DeclareMathOperator*{\VA}{\mathbb{V}ar}

\DeclareMathOperator*{\argmax}{argmax}

\newtheorem{theorem}{Theorem}

\begin{document}

\def\spacingset#1{\renewcommand{\baselinestretch}%
{#1}\small\normalsize} \spacingset{1}

\if0\blind
{
	{\title{\bf Active Learning of Piecewise Gaussian Process Surrogates}}

\author{Chiwoo Park$^{1}$, Robert Waelder$^{2}$, Bonggwon Kang$^{3}$, Benji Maruyama$^{2}$, Soondo Hong$^{3}$ and Robert B.~Gramacy$^{4}$ \\
\small{$^{1}$ Department of Industrial and Systems Engineering, University of Washington, Seattle WA 98195, USA.} 

\and \small{$^{2}$ Materials and Manufacturing Directorate, Air Force Research Laboratory, Wright-Patterson Air Force Base, Dayton, OH 45433, USA.} 

\and \small{$^{3}$ Department of Industrial Engineering, Pusan National University, Busan 46241, South Korea.} 

\and \small{$^{4}$  Department of Statistics, Virginia Tech, Blacksburg, VA 24061, USA.}}
\date{}
\maketitle
} \fi
\if1\blind
{
	\title{\bf Active Learning of Piecewise Gaussian Process Surrogates}
	\author{}
	\date{}
	\maketitle
} \fi

\vspace{-30pt}
\begin{abstract}
Active learning of Gaussian process (GP) surrogates has been useful for optimizing experimental designs for physical/computer simulation experiments, and for steering data acquisition schemes in machine learning. In this paper, we develop a method for active learning of piecewise, Jump GP surrogates.  Jump GPs are continuous within, but discontinuous across, regions of a design space, as required for applications spanning autonomous materials design, configuration of smart factory systems, and many others. Although our active learning heuristics are appropriated from strategies originally designed for ordinary GPs, we demonstrate that additionally accounting for model bias, as opposed to the usual model uncertainty, is essential in the Jump GP context. Toward that end, we develop an estimator for bias and variance of Jump GP models.  Illustrations, and evidence of the advantage of our proposed methods, are provided on a suite of synthetic benchmarks, and real-simulation experiments of varying complexity.
\end{abstract}

\noindent%

\spacingset{1.5} 

\section{Introduction} \label{sec:intro}
The main goal of machine learning is to create an autonomous computer system that can learn from data with minimal human intervention \citep{mitchell1997edition}. In many machine learning tasks, one can control the data acquisition process in order to select training examples that target specific goals. Active learning (AL) -- or sequential  design of experiments in classical statistical jargon -- is the study of how to select data toward optimizing a given learning objective \citep[e.g.,][]{cohn1996active, lam2008sequential}. Here we consider AL for piecewise continuous  Gaussian process (GP) surrogate models. 	

Our motivating application is surrogate modeling of engineering systems, to explore and understand overall system performance and ultimately to optimize their design. A particular focus is on engineering systems whose behaviors intermittently exhibit abrupt jumps or local discontinuities across regimes of a design space. Such ``jump system'' behaviors are found in many applications. For example, carbon nanotube yield from a chemical vapor deposition (CVD) process varies depending on many design variables \citep{Nikolaev2016}. Changes in dynamics are mostly gradual, but process yield can suddenly jump around, depending on chemical equilibrium conditions, from `no-growth' to `growth' regions. Such jump system behaviors are universal to many material and chemistry applications owing to several factors (i.e., phase changes, activation energy).   Jump behaviors are also prevalent in engineering systems operating near capacity. When a system runs below its capacity, performance exhibits little fluctuation.  However, performance can suddenly break down as the system is forced to run slightly over its capacity \citep{kang2022}.  

Suitable surrogate models for jump systems must accommodate piecewise continuous functional relationships, where disparate input--output dynamics can be learned (if data from the process exemplify them) in geographically distinct regions on input/configuration space.   Most existing surrogate modeling schemes make an assumption of stationarity, and are thus not well-suited to such processes. AL strategies paired with such surrogates are, consequently, sub-optimal for acquiring training examples in such settings. For example, Gaussian processes \citep[GPs;][]{williams2006gaussian} are perhaps the canonical choice for surrogate modeling of physical and computer experiments \citep{santner2018design}.  They are flexible, nonparametric, nonlinear, lend a degree of analytic tractability, and provide well-calibrated uncertainty quantification (UQ) without having to tune many unknown quantities.  
But the canonical, relative-distance-based kernels used with GPs result in stationary processes. Therefore,  AL schemes paired with GPs exhibit space-filling behavior, which may not be appropriate for certain applications.  Representative examples include ``Active Learning Cohn" \citep[ALC]{cohn1996active},  ``Active Learning-MacKay" \citep[ALM]{mckay2000comparison}, and ``Active Learning with Mutual Information" \citep[MI]{krause2008near, beck2016sequential}. Space-filling designs, and their sequential analogues, are inefficient when input--output dynamics change across regions of the input space.  Intuitively, we need a higher density of training examples in harder-to-model regions, and near boundaries where regime dynamics change.

Regime-changing dynamics are inherently non-stationary: both position and relative distance information (in the input configuration space) is required for effective modeling. Examples of non-stationary GP modeling strategies from the geospatial literature abound \citep{sampson1992nonparametric,schmidt2003bayesian,paciorek2006spatial}.  The trouble with these approaches is that they are too slow, in many cases demanding enormous computational resources in their own right, or limited to two input dimensions.  Recent developments in the machine learning literature around deep GPs \citep{damianou2013deep} represent a promising alternative.  Input dimensions can be larger, and fast inference is provided by doubly stochastic variational inference \citep{salimbeni2017doubly}.  However, such methods are data-hungry, requiring tens of thousands of training examples before they are competitive with conventional GP methods. We note that an ALC-type criterion has been developed for deep GPs \citep{sauer2021active}, making them less data-hungry, but slow Markov chain Monte Carlo (MCMC) inference remains a bottleneck.

A class of methods built around divide-and-conquer strategies can offer the best of both worlds -- computational thrift with modeling fidelity -- by simultaneously imposing statistical and computational independence. The best-known examples include treed GPs \citep{gramacy2008bayesian, taddy2011dynamic, malloy2014near, Konomi2014} and Voronoi tessellation-based GPs \citep{kim2005analyzing, heaton2017nonstationary, pope2021gaussian, luo2021bayesian}.  Partitioning facilitates non-stationarity almost trivially, by independently fitting different GPs in different parts of the input space.  Sequential design/AL criteria have been adapted to some of these divide-and-conquer surrogates.  ALM and ALC, for example, have been adapted for treed GPs \citep{gramacy2009adaptive, taddy2011dynamic}. However the axis-aligned nature of the treed GP is not flexible enough to handle the complex, nonlinear manifold of regime change exhibited of many real datasets.

\citet{park2022} introduced the Jump GP to address this limitation in domain-partitioning. This approach seeks  a local approximation to an otherwise potentially complex domain-partitioning and GP-modeling scheme. Crucially, direct inference for the Jump GP enjoys the same degree of analytic tractability as an ordinary, stationary GP. However, good AL strategies have not been studied for the Jump GP model, which is the main focus of this paper. We are inspired by related work in jump kernel regression \citep{doi:10.1080/24725854.2021.1988770}.  However, it is important to remark that the work focused on a more limited class of (non-GP) nonparametric regression models. 

In the context of the Jump GP for nonstationary surrogate modeling, we propose to extend conventional AL strategies to consider model bias in addition to the canonical variance-based heuristics. We show that considering bias is essential in a non-stationary modeling setting.  In particular, ordinary stationary GP surrogates may show significant biases at test locations near regime changes.  The Jump GP can help mitigate this bias, but it does not completely remove it.  AL strategies that don't incorporate estimates of bias are limited in their ability to improve sequential learning of a Jump GP. The major contribution of this paper is to estimate both bias and variance for Jump GPs and parlay these into novel AL strategies for nonstationary surrogate modeling. 

The remainder of the paper is outlined as follows.  In Section \ref{sec:model}, we review relevant topics around the Jump GP and AL. Section \ref{sec:method} develops bias and variance estimation for Jump GPs. Using those estimates, we develop four AL heuristics for Jump GPs in Section \ref{sec:AL}.  We illustrate the numerical evaluation using synthetic benchmarks and two real data cases in Section \ref{sec:numerical}.  A summary and discussion conclude the paper in Section \ref{sec:conc}. 

\section{Review} \label{sec:model}
Here we review components essential to framing our contribution: GP surrogates, AL, partition-based modeling and the Jump GP. 

\subsection{Stationary GP surrogates}
\label{sec:gp}

Let $\mathcal{X}$ denote a $d$-dimensional input configuration space. Consider estimating an unknown function $f: \mathcal{X} \rightarrow \mathbb{R}$ relating inputs $\V{x}_i \in \mathcal{X}$ to a noisy real-valued response variable $y_i \stackrel{\mathrm{iid}}{\sim} \mathcal{N}(f(\V{x}_i), \sigma^2)$ using training data, $\mathcal{D}_N = \{(\V{x}_i, y_i), \; i = 1, \ldots, N \}$.
In GP regression, a finite collection $\V{f}_N=(f_1,\dots, f_N)$ of $f(\V{x}_i) \equiv f_i$ values is modeled as a multivariate normal (MVN) random variable.  A common specification involves a constant, scalar mean $\mu$, and $N \times N$ covariance matrix $\V{C}_N$: $\V{f}_N \sim \mathcal{N}_N(\mu \mathbf{1}_N, \V{C}_N)$, where $\M{1}_N$ is an $N$-dimensional column vector of ones.

Rather than treating all $\mathcal{O}(N^2)$ values in $\V{C}_N$ as ``tunable parameters'', it is common to use a kernel $c(\V{x}_i, \V{x}_j; \V{\theta})$ defining correlations in terms of a small number of hyperparameters $\V{\theta}$.  Most kernel families \citep{wendland2004scattered} are decreasing functions of the geographic ``distance'' between its arguments $\V{x}_i$ and $\V{x}_j$.  An assumption of {\em stationarity} is common, whereby $c(\V{x}_i, \V{x}_j; \V{\theta}) \equiv c(\V{x}_i - \V{x}_j; \V{\theta})$, i.e., only relative displacement $\V{x}_i - \V{x}_j$ between inputs, not their positions, matters for modeling.  As discussed further, below, a stationarity assumption can be limiting and relaxing this is a major focus of the methodology introduced in this paper.  

Integrating out latent $\V{f}_N$ values, to obtain a distribution for $\V{y}_N$, is straightforward because both are Gaussian.  This leads to the marginal likelihood $\V{y}_N \sim \mathcal{N}_N(\mu \mathbf{1}_N, \sigma^2\M{I}_N + \V{C}_N)$ which can be used to learn hyperparameters, where $\M{I}_N$ is an $N$-dimensional identity matrix.   Maximum likelihood estimates (MLEs) for $\hat{\mu}$ and $\hat{\sigma}^2$ have closed forms conditional on $\V{\theta}$. E.g., \citet{mu2017sequential} provides $\hat{\mu} = \V{1}_N^T (\hat{\sigma}^2\M{I}_N +\V{C}_N)^{-1} \V{y}_N / \V{1}_N^T (\hat{\sigma}^2\M{I}_N +\V{C}_N)^{-1} \V{1}_N$.  Estimates $\hat{\V{\theta}}$ depend on the kernel and requires numerical methods \citep{williams2006gaussian,santner2018design,gramacy2020surrogates}.

Analytic tractability extends to prediction.  Basic MVN conditioning from a joint model of $\V{y}_N$ and an unknown testing output $Y(\V{x}_*)$ gives that $Y(\V{x}_*) \mid \V{y}_N$ is univariate Gaussian.  Below we quote the distribution for the latent function value $\hat{f}(\V{x}) \equiv f(\V{x}_*) \mid \V{y}_N$, which is of more direct interest in our setting.  This distribution is also Gaussian, with
\begin{equation} \label{eq:staGP}
	\begin{array}{l l}
		\mbox{mean: } 		&   \mu(\V{x}_*) = \hat{\mu} + \V{c}_N^T (\hat{\sigma}^2\M{I}_N +\V{C}_N)^{-1} (\V{y}_N - \hat{\mu} \V{1}_N), \mbox{ and } \\
		\mbox{variance: } 	&   s^2(\V{x}_*) = c(\V{x}_*, \V{x}_*; \hat{\V{\theta}}) - \V{c}_N^T (\hat{\sigma}^2\M{I}_N +\V{C}_N)^{-1} \V{c}_N,
	\end{array}
\end{equation}
where $\V{c}_N = [c(\V{x}_i, \V{x}_*;  \hat{\V{\theta}}): i = 1,\dots,N]$ is an $N \times 1$ vector of the covariance values between the training data and the test data point.  Observe that evaluating these prediction equations, like evaluating the MVN likelihood for hyperparameter inference, requires inverting the $N \times N$ matrix $\hat{\sigma}^2\M{I}_N +\V{C}_N$, which is computationally expensive.  

\subsection{Divide-and-conquer GP modeling}

Partitioned GP models \citep{gramacy2008bayesian, kim2005analyzing} generally,
and the Jump GP \citep{park2022} specifically, consider an $f$ that is piecewise
continuous
\begin{equation} \label{eq:regmodel}
	f(\V{x}) = \sum_{k=1}^K f_k(\V{x}) 1_{\mathcal{X}_k}(\V{x}),
\end{equation}
where $\mathcal{X}_1, \mathcal{X}_2, \ldots, \mathcal{X}_K$ are a partition of $\mathcal{X}$.  Above, $1_{\mathcal{X}_k}(\V{x})$ is an indicator function that determines whether $\V{x}$ belongs to region $\mathcal{X}_k$, and each $f_k(\V{x})$ is a continuous function that serves as a basis for the regression model on region $\mathcal{X}_k$. Although variations abound,  here we take each functional piece $f_k(\V{x})$ to be a stationary GP, as described in Section \ref{sec:gp}. 

Typically, each $f_k$ is presumed independent conditional on the partitioning mechanism.  This assumption is summarized below for easy referencing later.
\begin{equation} \label{eq:indep}
	\mbox{Independence:} \qquad f_k \mbox{ is independent of } f_j \mbox{ for } j \neq k.
\end{equation}
Consequently, all hyperparameters describing $f_k$ may be analogously indexed and are treated independently, e.g., $\mu_k$, $\sigma_k^2$ and $\V{\theta}_k$.  Generally speaking,  data within region $\mathcal{X}_k$ are used to learn these hyperparameters, via the
likelihood applied on the subset of data $\mathcal{D}_N$ whose $\V{x}$-locations reside in $\mathcal{X}_k$.  Although it is possible to allow novel kernels $c_k$ in each region,
it is common to fix a particular form (i.e., a family) for use throughout.  Only its hyperarmeters $\V{\theta}_k$ vary across regions, as in $c(\cdot, \cdot; \V{\theta}_k$).  Predicting with $\hat{f}(\V{x}_*)$, conditional on a partition and estimated hyperparameters, is simply a matter of following Eq.~(\ref{eq:regmodel}) with ``hats''.  That is, with $\hat{f}_k$ defined analogously to Eq.~(\ref{eq:staGP}), i.e., using only $y$-values exclusive to each region.  In practice, the sum over indicators in Eq.~(\ref{eq:regmodel}) is bypassed and one simply identifies the $\mathcal{X}_k$ to which $\V{x}_*$ belongs and uses the corresponding $\hat{f}_k$ directly.  

Popular, data-driven partitioning schemes leveraged by local GP models include Voronoi tessellation \citep{kim2005analyzing, heaton2017nonstationary, pope2021gaussian, luo2021bayesian} or recursive axis-aligned, tree-based partitioning \citep{gramacy2008bayesian, taddy2011dynamic, malloy2014near, Konomi2014}. These ``structures'', defining $K$, and within-partition hyperparameters $(\mu_k, \V{\theta}_k, \sigma^2)$ may be jointly learned, via posterior sampling (e.g., MCMC) or by maximizing marginal likelihoods.  In so doing, one is organically learning a degree of non-stationarity.  Independent GPs, via disparate independently learned hyperparameters, facilitate a position-dependent correlation structure.  Learning separate $\sigma_k^2$ in each region can also accommodate heteroskedasticity \citep{binois2018practical}.  Such divide-and-conquer techniques can additionally bring computational gains.

\subsection{Local GP modeling}

Although there are many example settings where such partition-based GP models excel,  their rigid structure can be a mismatch to many important real-data settings.  The Jump GP \citep[JGP;][]{park2022} is motivated by such applications.   The idea is best introduced through the lens of local, approximate GP modeling  \citep[LAGP;][]{gramacy2015local}.  For each test location $\V{x}_*$, select a small subset of training data nearby $\V{x}_*$:  $\mathcal{D}_n(\V{x}_*) = \{(\V{x}_{i,*}, y_{i,*}) \}_{i=1}^n \subset \mathcal{D}_N$. Then, fit a conventional, stationary GP model $\hat{f}_n(\V{x}_*)$ to the local data $\mathcal{D}_n(\V{x}_*)$.  This is fast, because $\mathcal{O}(n^3)$ is much better than $\mathcal{O}(N^3)$ when $n \ll N$, and massively parallelizable over many $\V{x}_* \in \mathcal{X}$ \citep{gramacy2014massively}.  It has a nice divide-and-conquer structure, but it is not a partition model (\ref{eq:regmodel}).  Nearby $\mathcal{D}_n(\V{x}_*')$ might have some, all, or no elements in common.  LAGP can furnish biased predictions \citep{park2022} because independence \eqref{eq:indep}  is violated: local data $\mathcal{D}_n(\V{x}_*')$ might mix training examples from regions of the input space exhibiting disparate input-output dynamics.

A JGP differs from basic LAGP modeling by selecting local data subsets in such a way as a partition (\ref{eq:regmodel}) is maintained and independence (\ref{eq:indep}) is enforced, so that bias is reduced.  Toward this end, the JGP introduces a latent, binary random variable $Z_{i,*} \in \{0, 1\}$ to express uncertainties on whether a local data point $\V{x}_{i,*}$ belongs to a region of the input exhibiting the same (stationary) input-output dynamics as the test location $\V{x}_*$, or not: $Z_{i,*} = 1$ if $\V{x}_{i,*}$ and $\V{x}_*$ belong to the same region, and $Z_{i,*}=0$ otherwise. Conditional on $Z_{i,*}$ values, $i=1,\dots,N$, we may partition the local data $\mathcal{D}_n$ into two groups: $\mathcal{D}_{*} = \{i \in \{1,\ldots, n\}: Z_{i,*} = 1 \}$ and $\mathcal{D}_o = \{1,\ldots, n\} \backslash \mathcal{D}_*$.  

Complete the specification by modeling $\mathcal{D}_*$ with a stationary GP with a constant mean $m_*$ and a stationary covariance function $c(\cdot, \cdot; \V{\theta}_*)$, $\mathcal{D}_o$ with dummy likelihood $p(y_{i,*} \mid Z_{i,*} = 0) \propto u$ for some constant, $u$, and assign a prior for the latent variable $Z_{i,*}$  via a sigmoid link $\pi(z) = 1/(1+\exp(-z))$ applied to an unknown partitioning function $g(\V{x}, \V{\omega})$, 
\begin{equation} \label{eq:latent}
	p(Z_{i,*} = 1|\V{x}_{i,*}, \V{\omega}) = \pi(g(\V{x}_{i,*}, \V{\omega})),   
\end{equation}
where $\V{\omega}$ is another hyperparameter. The form of the parametric partitioning function $g$ influences the boundary dividing $\mathcal{D}_{o}$ and $\mathcal{D}_{*}$. At the local level,  linear or quadratic $g$ serves a good local Taylor approximation to complex domain boundaries. For a detailed discussion, one may refer to \citet{park2022}. Here we take the linear form $g(\V{x}) = \V{\omega}^T [1, \V{x}]$.

Specifically, for $\V{Z} = (Z_{i,*})_{i=1}^n$, $\V{f}_* = (f_{i,*})_{i=1}^n$ and $\V{\Theta} = \{\V{\omega}, m_*, \V{\theta}_*, \sigma^2\}$, the JGP model may be summarized as follows: 
\begin{align*}
	p(\V{y}_{n} \mid \V{f}_*, \V{Z}, \V{\Theta}) &= \prod_{i=1}^n \mathcal{N}_1(y_{i,*}|f_{i,*}, \sigma^2)^{Z_{i,*}} u^{1-Z_{i,*}},\\
	p(\V{Z} \mid \V{\omega}) &= \prod_{i=1}^n \pi(g(\V{x}_{i,*}, \V{\omega}))^{Z_{i,*}} (1-\pi(g(\V{x}_{i,*}, \V{\omega})))^{1-Z_{i,*}}, \\
	p(\V{f}_* \mid m_*, \V{\theta}_*) &= \mathcal{N}_n(\V{f}_* \mid m_*\V{1}_n, \M{C}_{nn}),
\end{align*}
where $\V{y}_{n} = (y_{i,*})_{i=1}^n$ and $\V{C}_{nn}$ is a $n \times n$ matrix with $c(\V{x}_{i,*}, \V{x}_{j,*}; \V{\theta}_*)$ as its $(i,j)$th element.

Conditional on $\V{\Theta}$, prediction $\hat{f}(\V{x}_*)$ follows
the usual equations (\ref{eq:staGP}) using local data $\mathcal{D}_n(\V{x}_*)$.  A detailed presentation is delayed until Section \ref{sec:method}.  Inference for latent $\V{Z}$ may proceed by the expectation maximization (EM) algorithm. However, a difficulty arises because the joint posterior distribution of $\V{Z}$ and $\V{f}_*$ is not tractable, complicating the E-step. As a workaround, \citet{park2022} developed a \textit{classification EM} \citep[CEM;][]{bryant1978asymptotic} which replaces the E-step with a pointwise maximum {\em a posteriori} (MAP) estimation of $\V{Z}$.

\subsection{Active Learning for GPs}
Active learning (AL) attempts to sustain a virtuous cycle between data collection and model learning.  Begin with training data of size $N$, $\mathcal{D}_N = \{(\V{x}_i, y_i), i = 1,\ldots, N\}$, such as a space-filling Latin hypercube design \citep[LHD;][]{lin2015latin}. Then augment $\mathcal{D}_N$ with a new data point $(\V{x}_{N+1}, y_{N+1})$ chosen to optimize a criterion quantifying an important aspect or capability of the model, and repeat. Perhaps the canonical choice is mean square prediction error (MSPE), comprising of squared bias and variance \citep{hastie2009elements}. 

Many machine learning algorithms are equipped with proofs of unbiasedness of predictions under regularity conditions.  When training and testing data  jointly satisfy a stationarity assumption, the GP predictor (\ref{eq:staGP}) is unbiased, and so the MSPE is equal to $s^2(\V{x}_*)$.  Consequently, many AL leverage this quantity. For example, ALM maximizes it directly: $\V{x}_{N+1} = \argmax_{\V{x}_* \in \mathcal{X}} \; s^2(\V{x}_*).$  In repeated application, this ALM strategy can be shown to approximate a maximum entropy design \citep[][Section 6.3]{gramacy2020surrogates}.

An integrated mean squared prediction error (IMSPE) criterion considers how the MSPE of GP is affected, globally in the input space, {\em after} injecting new data at $\V{x}_{N+1}$.  Let $s^2(\V{x}_{*}; \V{x}_{N+1})$ denote the predictive variance \eqref{eq:staGP} at a test location $\V{x}_*$, when the training data $\mathcal{D}_N$ is augmented with one additional input location $\V{x}_{N+1}$:
\begin{equation*}
	s^2(\V{x}_{*}; \V{x}_{N+1}) = c(\V{x}_*, \V{x}_*; \hat{\V{\theta}}) - \V{c}_{N+1}^T (\hat{\sigma}^2\M{I}_{N+1} +\V{C}_{N+1})^{-1} \V{c}_{N+1},
\end{equation*}
where $\V{c}_{N+1} = (c(\V{x}_i, \V{x}_*; \hat{\V{\theta}}))_{i=1}^{N+1}$, and $\V{C}_{N+1}$ has $c(\V{x}_i, \V{x}_j; \hat{\V{\theta}})$ as its $(i+j)^\mathrm{th}$ element. 
The criterion is $
	\mbox{IMSPE}(\V{x}_{N+1}) = \int_{\mathcal{X}}	s^2(\V{x}_{*}; \V{x}_{N+1}) d\V{x}_*,$
which has a closed form \citep{binois2018replication}, although in machine learning quadrature-based ALC is preferred \citep{seo2000gaussian}.  

Such variance-only criteria make sense when data satisfies the unbiasedness condition, i.e., under stationarity, which can be egregiously violated in
many real-world settings.  In Bayesian optimization contexts, acquisition criteria have been extended to account for this bias \citep{lam2008sequential, mu2017sequential}, but we are not aware of any analogous work for AL targeting overall accuracy.   

\section{Bias--variance decomposition for JGPs} 
\label{sec:method}

The following discussion centers around predictive equations for a JGP: essentially Eq.~(\ref{eq:staGP}) with $\mathcal{D}_n(\V{x}_*)$.  For convenience,
these are re-written here, explicitly in that JGP notation.   Let $\hat{Z}_i$ represent the MAP estimate at convergence (of the CEM algorithm) and let $\mathcal{D}_{n,*} = \{i \in \{1,\ldots, n\}: \hat{Z}_i = 1\}$ denote the estimate of ${\mathcal{D}}_*$ with $n_*$ being the number of training data pairs in the set. Let $\V{y}_* = (y_i)_{i \in \mathcal{D}_{n,*}}$ is a $n_* \times 1$ vector of selected local data. Given the GP prior assumed on $\mathcal{D}_*$ (in JGP) and the parameter estimate $\hat{\V{\Theta}}$, $f_*(\V{x}_*)$ and $\V{y}_{*}$ follows a MVN, 
\begin{equation*} 
	\left[ \begin{array}{c}  f_*(\V{x}_*) \\ \V{y}_{*} \end{array} \right] \sim \mathcal{N}_{n_*+1}\left(
	\left[ \begin{array}{c}  \hat{m}_* \\  \hat{m}_* \V{1}_{n_*} \end{array} \right], 
	\left[ \begin{array}{c c} c(\V{x}_*, \V{x}_*; \hat{\V{\theta}}_*)  & \V{c}_{*}^T\\  \V{c}_{*} & \hat{\sigma}^2\M{I}_{n_*} +\V{C}_{**} \end{array} \right]
	\right),
\end{equation*}
where $\V{c}_{*} = (c(\V{x}_{i,*}, \V{x}_*; \hat{\V{\theta}}_*))_{i \in \mathcal{D}_{n,*}}$ is a column vector of the covariance values between $\V{y}_*$ and $f(\V{x}_*)$, and $\V{C}_{**} = (c(\V{x}_i, \V{x}_j; \hat{\V{\theta}}_*))_{i, j \in \mathcal{D}_{n,*}}$ is a square matrix of covariances evaluated for all pairs of $\V{y}_*$. Here, $\hat{\sigma}^2$ and $\hat{\V{\theta}}_*$ represent the MLEs of $\sigma^2$ and $\V{\theta}_*$ respectively, and $\hat{m}_*$ is the MLE of $m_*$, which
has the form
\begin{equation}
	\label{eq:mstar}
	\hat{m}_* = \frac{\V{1}_{n_*}^T (\hat{\sigma}^2\M{I}_{n_*} +\V{C}_{**})^{-1} \V{y}_*}{\V{1}_{n_*}^T (\hat{\sigma}^2\M{I}_{n_*} +\V{C}_{**})^{-1} \V{1}_{n_*}}.
\end{equation}

The posterior predictive distribution of $\hat{f}_*(\V{x})$ at a test location $\V{x}_*$ is univariate Gaussian with
\begin{equation} \label{eq:localGP4}
	\begin{array}{l l}
		\mbox{mean: } 		&   \mu_J(\V{x}_*) = \hat{m}_* + \V{c}_{*}^T (\hat{\sigma}^2\M{I}_{n_*} +\V{C}_{**})^{-1} (\V{y}_{*} - \hat{m}_* \V{1}_{n_*}), \mbox{ and } \\
		\mbox{variance: } 	&   s^2_J(\V{x}_*) = c(\V{x}_*, \V{x}_*; \hat{\V{\theta}}_*) - \V{c}_{*}^T (\hat{\sigma}^2\M{I}_{n_*} +\V{C}_{**})^{-1} \V{c}_{*},
	\end{array}
\end{equation}
Subsections which follow break down the mean $\mu_J(\V{x}_*)$ and variance $s_J^2(\V{x}_*)$ quoted in Eq.~(\ref{eq:localGP4}), in terms of their contribution to bias and variance of a JGP predictor, respectively, with an eye toward AL application in Section \ref{sec:AL} as an estimator of MSPE:
\begin{equation}
	\mathrm{MSPE}[\mu_J(\V{x}_*)] = \mathrm{Bias}^2[\mu_J(\V{x}_*)] + 	\VA[\mu_J(\V{x}_*)].  \label{eq:mspe}
\end{equation}

\subsection{Bias}
The following theorem gives the bias of $\mu_J(\V{x}_*)$ under an assumption that holds for test points where at most two regions are separated by a border. The proof is in Appendix A. 
\begin{theorem}
	In estimating an unknown function $f$ in the form of \eqref{eq:regmodel}, assume that the local data group $\mathcal{D}_o$ belongs to one of the $K$ regions $\mathcal{X}_1,\ldots, \mathcal{X}_K$.  This corresponds to the case that local data $\mathcal{D}_n(\V{x}_*)$ comes from two regions with each of $\mathcal{D}_*$ and $\mathcal{D}_o$ belonging to one of the $K$ regions. Let $Z_*$ denote the latent binary variable representing the risk that the test point $\V{x}_*$ is mis-classified into $D_o$, where the probability of mis-classification is  given by $P(Z_*=0) = 1-\pi(g(\V{x}_*, \V{\omega}))$, according to the prior \eqref{eq:latent}.
	The bias of $\mu_J(\V{x}_*)$ is 
	\begin{equation} \label{eq:bias0}
		\begin{split}
			\mathrm{Bias}[\mu_J(\V{x}_*)] & =  (m_* - m_o) \sum_{j \in \mathcal{D}_{n,*}} \alpha_{j} \{(1-p_j) p_* - p_j (1-p_*) \} \\
			& \quad + (m_* - m_o) \sum_{i \in \mathcal{D}_{n,*}}\sum_{j \in \mathcal{D}_{n,*}} \alpha_{j}\beta_{i} \{  p_j (1-p_i) - (1-p_j) p_i  \},
		\end{split}
	\end{equation}
	where $m_o = \EX[f(\V{x}_*)|Z_*=0]$, $\alpha_{i}$ is the $i^\mathrm{th}$ element of $\V{\alpha} = \V{1}_{n_*}^T (\sigma^2\M{I}_{n_*} +\V{C}_{**})^{-1} / (\V{1}_{n_*}^T (\sigma^2\M{I}_{n_*} +\V{C}_{**})^{-1} \V{1}_{n_*}),$ $p_* = p(Z_*=1)$, $p_i = P(Z_{i,*}=1)$, and $\beta_{j}$ is the $j^\mathrm{th}$ element of $(\sigma^2\M{I}_{n_*} +\V{C}_{**})^{-1}\V{c}_{*}$. 
\end{theorem}

Evaluating this expression is complicated by many unknown quantities such as $m_*$, $m_o$, $p_*$ and $p_i$. We develop a plug-in estimate here and provide a numerical scheme in Appendix~B. 

The quantity $m_*$ may be estimated by a local mean estimate $\hat{m}_*$. Similarly, with a uniform likelihood for $\mathcal{D}_*$, $m_o$ may be estimated via sample mean, $$\hat{m}_o = \frac{1}{n -n_*}\sum_{i \in \mathcal{D}_n(\V{x}_*) \backslash \mathcal{D}_{n, *}} y_{i,*}.$$  Probabilities $p_j$ and $p_*$ may be estimated via Eq.~\eqref{eq:latent} with $\V{\omega}$ estimated by the JGP.  Let $\hat{p}_j$ and $\hat{p}_*$ denote those estimates. Inserting them into \eqref{eq:bias0} yields the following plug-in estimate:
\begin{equation} \label{eq:bias4}
	\begin{split}
		\widehat{\mathrm{Bias}}[\mu_J(\V{x}_*)] &= (\hat{m}_* - \hat{m}_o)\sum_{j \in \mathcal{D}_{n,*}} \alpha_{j} \{(1-\hat{p}_j) \hat{p}_* - \hat{p}_j (1-\hat{p}_*) \} \\
		& \quad + (\hat{m}_* - \hat{m}_o) \sum_{i \in \mathcal{D}_{n,*}}\sum_{j \in \mathcal{D}_{n,*}} \alpha_{j}\beta_{i} \{ \hat{p}_j (1-\hat{p}_i)  - (1-\hat{p}_j) \hat{p}_i\}.
	\end{split}
\end{equation}	

It is worth remarking that $\widehat{\mathrm{Bias}}[\mu_J(\V{x}_*)]$ in Eq.~(\ref{eq:bias4}) is influenced by accuracy of $\V{\hat{Z}}$, or in other words by the classification accuracy of local data furnished by the CEM algorithm. The first term in $\widehat{\mathrm{Bias}}[\mu_J(\V{x}_*)]$ requires the summation of more non-zero terms as the probability that $Z_{j,*} \neq Z_*$ increases (for the selected data $\V{y}_*$), i.e, when the selected data have low probabilities of being from the region of a test location. Conversely, the second term in $\widehat{\mathrm{Bias}}[\V{x}_*]$ would require summing more non-zero terms as the total probabilities of the selected data $\V{y}_*$ (being from heterogeneous regions) increases, i.e., when the selected data are highly likely from heterogeneous regions. Some visuals will be provided momentarily in Section \ref{sec:illus} along with a comprehensive illustration.  But first, we wish to complete the MSPE decomposition (\ref{eq:mspe}) with an estimate of predictive variance. 

\subsection{Variance}
\label{sec:var}

Conditional on $\V{Z}$, the variance of $\mu_J(\V{x}_*)$ is given by $s_J^2(\V{x}_*)$ in Eq.~(\ref{eq:localGP4}). To make this dependency explicit, we rewrite this variance as $s_J^2(\V{x}_*; \V{Z})$. The law of total probability can be used to obtain the overall variance of $\mu_J(\V{x}_*)$, unconditional on $\V{Z}$:
\begin{equation} 
	\VA[\mu_J(\V{x}_*)] = \sum_{\V{Z} \in \mathbb{B}^n} s_J^2(\V{x}_*; \V{Z}) p(\V{Z}),  \label{eq:fullsum}
\end{equation}
where $\mathbb{B}^n$ represents the set of all $n$-dimensional binary vectors. Evaluating this expression (\ref{eq:fullsum}) in practice is cumbersome as the number of distinct settings of $\V{Z}$ grows as $2^N$. To streamline the evaluation, we prefer to short-circuit an exhaustive enumeration by bypassing {\em a posteriori} improbable settings, instead considering only $\V{Z}$'s with high $p(\V{Z})$ values. Toward that, let $\mathcal{B}_{r}(\V{\hat{Z}}) = \{ \V{Z} \in \mathbb{B}^n: Z_{j,*} = \hat{Z}_{j,*} \mbox{ except for } r \mbox{ elements} \}.$ Since $\mathbb{B}^n = \bigcup_{r=0}^{n} \mathcal{B}_{r}(\V{\hat{Z}})$, and $\sum_{r=0}^n \sum_{\V{Z} \in \mathcal{B}_{ r}(\V{\hat{Z}})} p(\V{Z}) = 1$, we have
\begin{equation}
	\VA[\mu_J(\V{x}_*)] \approx \frac{\sum_{r=0}^n \sum_{\V{Z} \in \mathcal{B}_{r}(\V{\hat{Z}})} s^2_J(\V{x}_*; \V{Z}) p(\V{Z})}{ \sum_{r=0}^n \sum_{\V{Z} \in \mathcal{B}_{r}(\V{\hat{Z}})} p(\V{Z}) },
\end{equation} 
where $p(\V{Z})$ can be estimated as $\hat{p}(\V{Z}) = \prod_{i=1}^n \hat{p}_i^{Z_{i,*}} (1-\hat{p}_i)^{1-Z_{i,*}}$. 

Due to the nature of CEM inference, estimated $\hat{p}_i$ are highly concentrated around 0 and 1, which we demonstrate empirically in Appendix C. We exploit this highly concentrated distribution of $\hat{p}_i$ to aggressively short-circuit an exhaustive sum (\ref{eq:fullsum}).  In so doing, we obtain a reasonable approximation because it can be shown that $\hat{p}(\V{Z})$ quickly goes to  zero for $\V{Z} \in \mathcal{B}_{ r}(\V{\hat{Z}})$ as $r$ increases.  For example, suppose $r=1$. For $\V{Z} \in \mathcal{B}_{1}(\V{\hat{Z}})$, let $j$ denote the index of an element with $Z_{j,*} \neq \hat{Z}_{j,*}$, and assume that $\min(\hat{p}_j, 1-\hat{p}_j)$ within the 95$^\mathrm{th}$ percentile, which is empirically less than 0.1 [Appendix C]. Then, we have 
	$$\hat{p}(\V{Z}) = \hat{p}(\V{\hat{Z}}) \frac{\min(\hat{p}_j, 1-\hat{p}_{j})}{\max(\hat{p}_j, 1-\hat{p}_j)} \le \frac{0.1}{1-0.1},$$
because $\hat{p}(\V{\hat{Z}}) \le 1$ and $\min(\hat{p}_j, 1-\hat{p}_{j})/\max(\hat{p}_j, 1-\hat{p}_j) < 0.1 / (1-0.1)$. The same holds more generally, for larger $r$, where $\V{Z} \in \mathcal{B}_{r}(\V{\hat{Z}})$:
\begin{equation*}
\hat{p}(\V{Z}) \le \left(\frac{0.1}{1-0.1}\right)^r.
\end{equation*}	
This value quickly decreases as $r$ increases.

Since $\hat{p}(\V{Z})$ decreases exponentially as $r$ increases, we can obtain a good approximation to the exhaustive variance (\ref{eq:fullsum}) by the truncated series,
\begin{equation} \label{eq:var}
	\widehat{\VA}_{R}[\mu_J(\V{x}_*)] =  \frac{\sum_{r=0}^{R} \sum_{\V{Z} \in \mathcal{B}_{r}(\V{\hat{Z}})} s_J(\V{x}_*; \V{Z}) \hat{p}(\V{Z})}{ \sum_{r=0}^{R}  \sum_{\V{Z} \in \mathcal{B}_{r}(\V{\hat{Z}})} \hat{p}(\V{Z}) }.
\end{equation}
This expression (\ref{eq:var}) approaches the true $\VA[\mu_J(\V{x}_*)]$ as $R \rightarrow n$, requiring the evaluation of $\sum_{r=0}^{R} {n \choose r}$ terms. When $R = 0$, the approximation in Eq.~(\ref{eq:var}) reduces to $s^2_J(\V{x}_*; \V{\hat{Z}})$, which is a popular acquisition criteria in its own right, forming the basis of ALM. Larger $R$-values yield only marginal gains, both in terms of the magnitude of the resulting variance estimate and progress towards the true calculation. Again, looking ahead to our illustration coming next, our variance estimates (\ref{eq:var}) using $R=0$ are within ten percent of the truth (\ref{eq:fullsum}), specifically 9.0181 and around 10, respectively.  
The former improves to 9.0194 with $R=1$. We prefer $R=0$ for our empirical work later, although it's certainly easy to try modestly larger $R$-values.  Other implementation details are deferred to Section \ref{sec:numerical}.

\subsection{Illustration}
\label{sec:illus}
\begin{figure}[ht!]
	\centering
	\includegraphics[width=0.8\textwidth]{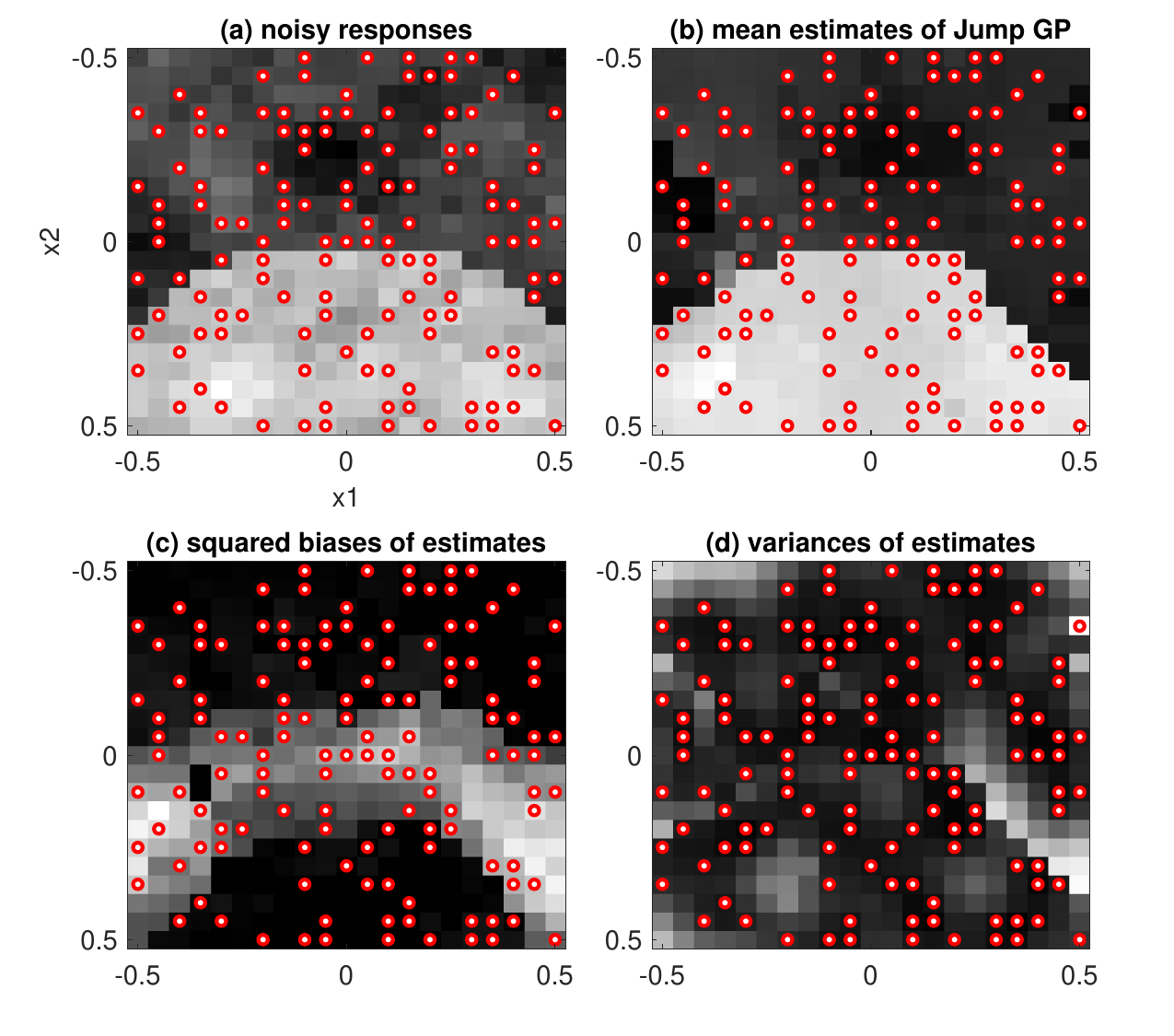}
	\vspace{-30pt}
	\caption{Bias and variances of JGP for a test function over a 2d grid (a).  Red dots in all panels represent the training inputs. Panel (b) shows mean estimates from JGP with those data. Panels (c) and (d) show $\widehat{\mathrm{Bias}}[\mu_J(\V{x}_*)]$ and $\widehat{\VA}_{0}[\mu_J(\V{x}_*)]$ of the JGP estimates. In all panels, lighter colors represent higher values.} \label{fig12}
\end{figure}

Here we use a 2d toy example with a curvy boundary to illustrate how the bias and variance of JGP estimates may be decomposed via the approximations laid out above. See Figure \ref{fig12} (a). The response function for each region is randomly drawn from an independent GP with distinct constant mean $\mu_k \in \{0, 27\}$ and a squared exponential covariance function, $
		c(\V{x}, \V{x}'; \V{\theta}_k)  = 9\exp\left\{  - \frac{1}{200} (\V{x}-\V{x}')^T (\V{x}-\V{x}') \right\}$, and independent $\mathcal{N}(0, 2^2)$ noise.

Figure \ref{fig12} (a) shows a realization of one such surface.  Overlaid are $N=132$ training inputs selected at random from a $21 \times 21$ grid over the input domain. Noisy responses at those training inputs are used to estimate the JGP. Figure \ref{fig12} (b) shows JGP-estimated predictive means over that grid. Figure \ref{fig12} (c) and (d) show the decomposition of MSPE (\ref{eq:mspe}) via calculated values of $\widehat{\mathrm{Bias}}[\mu_J(\V{x}_*)]$ and $\widehat{\VA}_{0}[\mu_J(\V{x}_*)]$, respectively. Observe that high bias clusters around the boundary of the two regions, whereas high variance clusters around parts of the input space where training data are sparse.  The former phenomenon is specific to the JGP setting and represents a feature we aim to exploit for AL. For JGP learning, you want more data where bias is high. The latter is a classic characteristic of (otherwise stationary) GP modeling; more data where training examples are sparse.

\section{Active Learning for Jump GPs} \label{sec:AL}

Given training data $\mathcal{D}_N = \{(\V{x}_i, y_i), i = 1,\ldots, N\}$, we consider an AL setup that optimizes the acquisition of new training data by selecting a tuple of input coordinates $\V{x}_{N+1}$ among candidate positions in $\mathcal{X}_C$, of size $N_C$, according to some criterion.  For all  numerical examples in this paper, we take $N_C = 100d$ and form $\mathcal{X}_C$ via maximin LHD \citep{santner2018design}.	
Once $\V{x}_{N+1}$ has been selected, it is run to obtain $y_{n+1} = f(\V{x}_{n+1}) + \varepsilon$. The training data is augmented to form $\mathcal{D}_{N+1} = \mathcal{D}_N \cup (\V{x}_{N+1}, y_{N+1})$. Finally, the model is refit and the process repeats with $N \leftarrow N+1$.  Here we develop three AL criteria for JGPs that are designed to account for predictive model bias and variance to varying degree.

\subsection{Acquisition functions}

First consider an ALM-type criterion, selecting $\V{x}_{N+1}$ where the MSPE in Eq. (\ref{eq:mspe}) is maximized.  Exploiting bias \eqref{eq:bias4} and variance  \eqref{eq:var} estimates, we select $\V{x}_{N+1} \in \mathcal{X}_C$ as 
\begin{equation}
	\begin{split}
		\V{x}_{N+1} = \argmax_{\V{x}_{*} \in \mathcal{X}_C} \quad  \widehat{\mathrm{MSPE}}(\V{x}_*) =  \widehat{\mathrm{Bias}}[\mu_J(\V{x}_*)]^2  + \widehat{\VA}[\mu_J(\V{x}_*)].
	\end{split} 
\end{equation}
We refer to this AL critera as \textit{Maximum MSPE Acquisition}. 

Second, we explore an IMSPE (or ALC) type criterion that sequentially selects new data points among candidate locations in order to maximize the amount by which  a new acquisition would reduce total variance throughout the input space. To evaluate how the MSPE of a JGP changes with a new data $(\V{x}_{N+1}, y_{N+1})$, we must first understand how the addition of new training data affects the $n$-nearest neighbors of a test location $\V{x}_*$. 

Let $R(\V{x}_*) = \max_{\V{x}_i \in D_n(\V{x}_*)} d(\V{x}_{*}, \V{x}_i)$ denote the size of the neighborhood $D_n(\V{x}_*)$ before the new data is added, where $d(\cdot, \cdot)$ is Euclidean distance in $\mathcal{X}$ space. When $d(\V{x}_{N+1}, \V{x}_*) \ge R(\V{x}_*)$, the neighborhood does not change with the injection of the new data. Therefore, the change in MSPE would be zero at $\V{x}_*$. Consequently, we  only consider test locations $\V{x}_*$ satisfying $d(\V{x}_{N+1}, \V{x}_*) < R(\V{x}_*)$ going forward. Let $\mathcal{X}(\V{x}_{N+1})$ represent all test locations satisfying that condition.  For $\V{x}_* \in   \mathcal{X}(\V{x}_{N+1})$, let $\mathcal{D}_{n}(\V{x}_*, \V{x}_{N+1})$ represent the new $n$-nearest neighborhood of $\V{x}_*$. Without loss of generality, consider $\V{x}_{n,*} = \argmax_{\V{x}_{i,*} \in D_n(\V{x}_*)} ||\V{x}_{*} - \V{x}_{i,*} ||$. Then, we can write $
	\mathcal{D}_n(\V{x}_*; \V{x}_{N+1}) = \mathcal{D}_n(\V{x}_*) \cup \{(\V{x}_{N+1}, y_{N+1}) \} - \{(\V{x}_{n,*}, y_{n,*}) \}.$

When $(\V{x}_{N+1}, y_{N+1})$ are known, one can fit a JGP to $\mathcal{D}_n(\V{x}_*; \V{x}_{N+1})$. Let $\mu_J(\V{x}_*| \V{x}_{N+1},\V{y}_{N+1})$ and $s_J(\V{x}_*| \V{x}_{N+1},\V{y}_{N+1})$ denote the posterior mean and variance, based on \eqref{eq:staGP}. The corresponding MSPE can be achieved using \eqref{eq:bias4} and \eqref{eq:var}. The corresponding MSPE is 
\begin{equation}
	\begin{split}
		\widehat{\mathrm{MSPE}}(\V{x}_* \mid \V{x}_{N+1},y_{N+1}) = \widehat{\mathrm{Bias}}[\mu_J(\V{x}_* \mid \V{x}_{N+1},y_{N+1})]^2 + \widehat{\VA}[\mu_J(\V{x}_*| \V{x}_{N+1},y_{N+1})].
	\end{split} 
\end{equation}
In an AL setting, however,  $y_{N+1}$ is unknown at the time that an acquisition decision is being made.  To assess the potential value of it's input, $\V{x}_{N+1}$, we propose to estimate $y_{N+1}$, via the  posterior (predictive) distribution of $y_{N+1}$ based on the original data around $\V{x}_{N+1}$, i.e., $\mathcal{D}_n(\V{x}_{N+1})$. Specifically, for a JGP $Y(\V{x}_{N+1}) \mid \mathcal{D}_n(\V{x}_{N+1})$ via Eq.~\eqref{eq:localGP4} we have $p(y_{N+1} \mid \mathcal{D}_n(\V{x}_{N+1})) \equiv \mathcal{N}_1(\mu_J(\V{x}_{N+1}), \sigma^2 + s^2_J(\V{x}_{N+1}))$.  

IMSPE may then be defined as the average MSPE over $\V{x}_*$ and $y_{N+1}$,
\begin{equation}
	\widehat{\mathrm{IMSPE}}(\V{x}_{N+1}) = \int \int \widehat{\mathrm{MSPE}}(\V{x}_* \mid \V{x}_{N+1},\mu_J(\V{x}_{N+1})) p(y_{N+1} \mid \mathcal{D}_n(\V{x}_{N+1})) d\V{x}_* dy_{N+1}.
\end{equation}
For acquisition, we are primarily interested in how injecting $\V{x}_{N+1}$ into the design improves the IMSPE, which may be measured as
$\Delta \widehat{\mathrm{IMSPE}}(\V{x}_{N+1}) = \int \Delta \widehat{\mathrm{MSPE}}(\V{x}_*\mid \V{x}_{N+1}) d\V{x}_*
$,
where the integrand represents the improvement of MSPE at $\V{x}_*$, $$\Delta \widehat{\mathrm{MSPE}}(\V{x}_* \mid \V{x}_{N+1}) = \widehat{\mathrm{MSPE}}(\V{x}_*) - \int \widehat{\mathrm{MSPE}}(\V{x}_* \mid \V{x}_{N+1},y_{N+1}) p(y_{N+1} \mid \mathcal{D}_n(\V{x}_{N+1})) dy_{N+1}.$$
Note that the term $\Delta \widehat{\mathrm{MSPE}}(\V{x}_* \mid \V{x}_{N+1})$ is non-zero only for $\V{x}_*$ satisfying $d(\V{x}_{N+1}, \V{x}_*) < R(\V{x}_*)$. Let $\mathcal{X}(\V{x}_{N+1}) = \{\V{x}_{*} \in \mathcal{X}, d(\V{x}_{N+1}, \V{x}_*) < R(\V{x}_*)\}$ represent the set of such test locations. The improvement can be further refined to 
\begin{equation*}
	\Delta \widehat{\mathrm{IMSPE}}(\V{x}_{N+1}) = \int_{\mathcal{X}(\V{x}_{N+1})} \Delta \widehat{\mathrm{MSPE}}(\V{x}_* \mid \V{x}_{N+1}) d\V{x}_* .
\end{equation*}
Unfortunately, this integral is not available in a closed form, but we have found that it can be approximated accurately using Monte Carlo-based quadrature.  First draw a finite number of $\V{x}_*$'s, based on a minimax distance design or other space filling design. The size of the draws is denoted by $N_*$, which depends on the size of domain $\mathcal{X}$. We used $N_* = 20 \times d$ for all synthetic examples with $\mathcal{X} = [-0.5, 0.5]^d$. For each $\V{x}_* \in \mathcal{X}(\V{x}_{N+1})$,  draw i.i.d.~samples of $y_{N+1}$ from $p(y_{N+1} \mid \mathcal{D}_n(\V{x}_{N+1}))$ and refit the JGP for each draw of $y_{N+1}$ to evaluate the integral in $\Delta \widehat{\mathrm{MSPE}}(\V{x}_*|\V{x}_{N+1})$. In the actual implementation, we use one deterministic draw of $y_{N+1}$ at the posterior predictive mean, $\mu_J(\V{x}_{N+1})$, as a computational shortcut.  Finally, repeat to accumulate averages of $\Delta \widehat{\mathrm{MSPE}}(\V{x}_* \mid \V{x}_{N+1})$ to approximate $\Delta \widehat{\mathrm{IMSPE}}(\V{x}_{N+1})$.   Selecting $\V{x}_{N+1} = \argmax_{\V{x} \in \mathcal{X}_C} \Delta  \widehat{\mathrm{IMSPE}}(\V{x})$ in this way yields what we term a \textit{Minimum IMSPE Acquisition}, as maximizing the improvement $\Delta  \widehat{\mathrm{IMSPE}}(\V{x})$ corresponds to minimizing the IMSPE.

Our final AL criteria is \textit{Maximum Variance Acquisition}
\begin{equation}
	\V{x}_{N+1} = \argmax_{\V{x}_* \in \mathcal{X}_C} \quad  \widehat{\VA}[\mu_J(\V{x}_*)].
\end{equation}
Like the first criteria, above, this is an ALM-type, but one which ignores predictive bias.  Focusing exclusively on variance is known to produce space-filling designs.  Consequently, it serves as a sensible benchmark competitor.

\subsection{Illustration} \label{sec:AL_illus}

Here we use a simple toy example to explore the three acquisition functions presented in the previous section. For effective visualization, we use a two-dimensional rectangular domain $[-0.5, 0.5]^2$. 
\begin{figure}[ht!]
	\centering
	\includegraphics[width=0.8\textwidth]{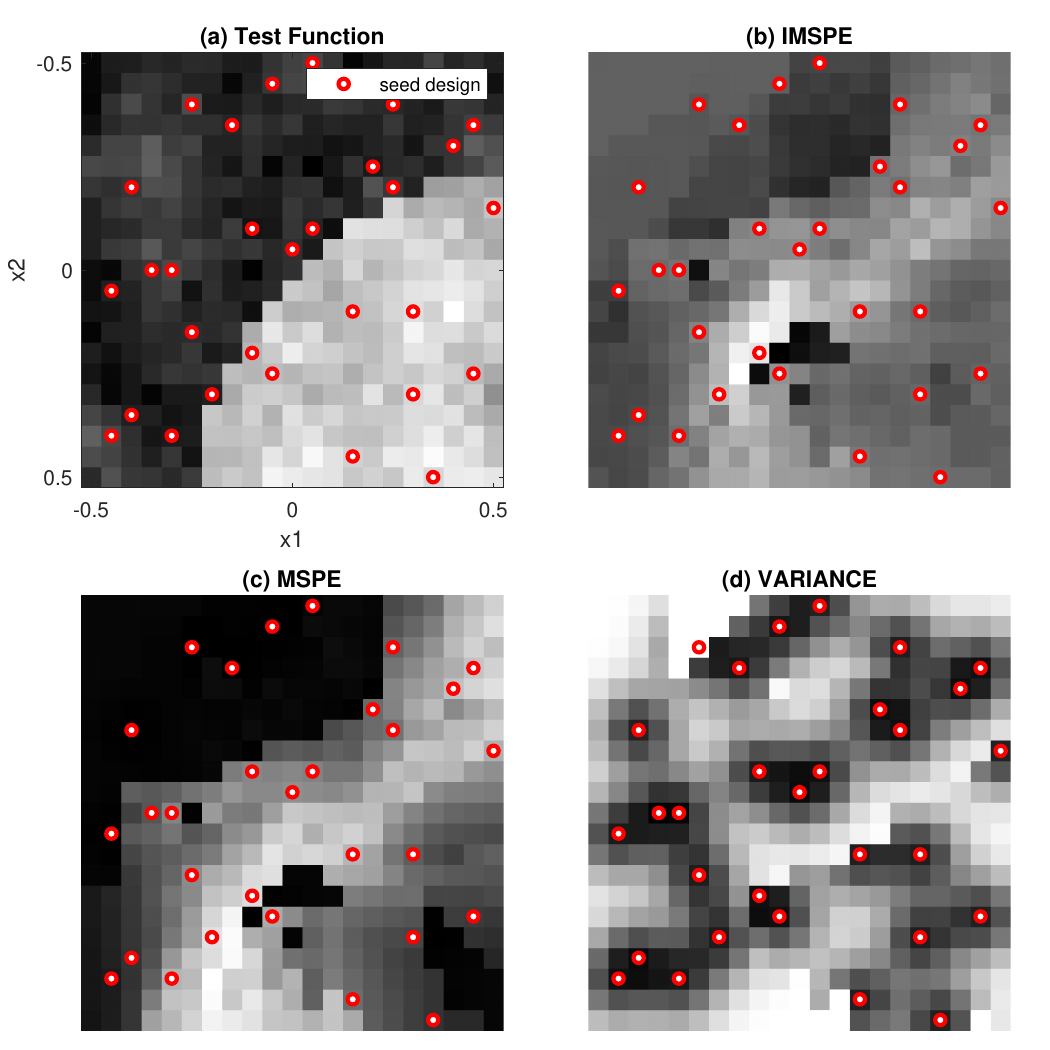}
	\vspace{-20pt}
	\caption{Three acquisition functions evaluated over $21 \times 21$ grid on $[-0.5, 0.5]^2$.  Panel (a) shows the data. Panels (b--d) show $\Delta \widehat{\mathrm{IMSPE}}(\V{x}_{*})$, $\widehat{\mathrm{MSPE}}(\V{x}_*)$ and $\widehat{\VA}[\mu_J(\V{x}_*)]$. The background displays acquisition function values with lighter colors indicating higher values.} \label{fig15}
\end{figure}
In Figure \ref{fig15} (a) this domain is partitioned into two regions, $\mathcal{X}_1$ (lighter shading) and $\mathcal{X}_2$ (darker).  The noisy response function for each region is randomly drawn from an independent GP with the same regional means and covariance functions used in our Section \ref{sec:illus} illustration. AL
is initialized with thirty seed-data positions via LHD, and these are overlayed onto all panels of the figure. Given noisy observations at these seed positions, we evaluated our three AL criteria on a $21 \times 21$ grid of candidates $\mathcal{X}_C$ in the input domain. We use a dense $\mathcal{X}_C$ here to aid with visualization. Panels \ref{fig15} (b--d) provide visuals for their acquisition in greyscale. For example, the maximum variance criterion in (d) indicates $\widehat{\VA}[\mu_J(\V{x}_*)]$. Observe that this criteria is inversely proportional to the positions (and densities, locally) of the seed data.  It exhibits behavior similar to conventional variance-based criteria (such as ALM or ALC) derived from stationary GPs. One slight difference, however, is that that the variance of the JGP is slightly higher around regional boundaries, compared to an ordinary stationary GP. Around the boundaries, local data is bisected, and only one section is used for JGP prediction, which makes the variance elevated there. Later, in Figure \ref{fig13}, we will see that slightly more data positions are selected around regional boundaries with the variance criterion. As shown in panels (b) and (c), the IMSPE and MSPE values consider  model bias and variance. Around regional boundaries,  bias values dominate variance estimates. Therefore, as we will show in Figure \ref{fig13}, data acquisitions are highly concentrated around regional boundaries.

\begin{figure}[ht!]
	\centering
	\includegraphics[width=\textwidth]{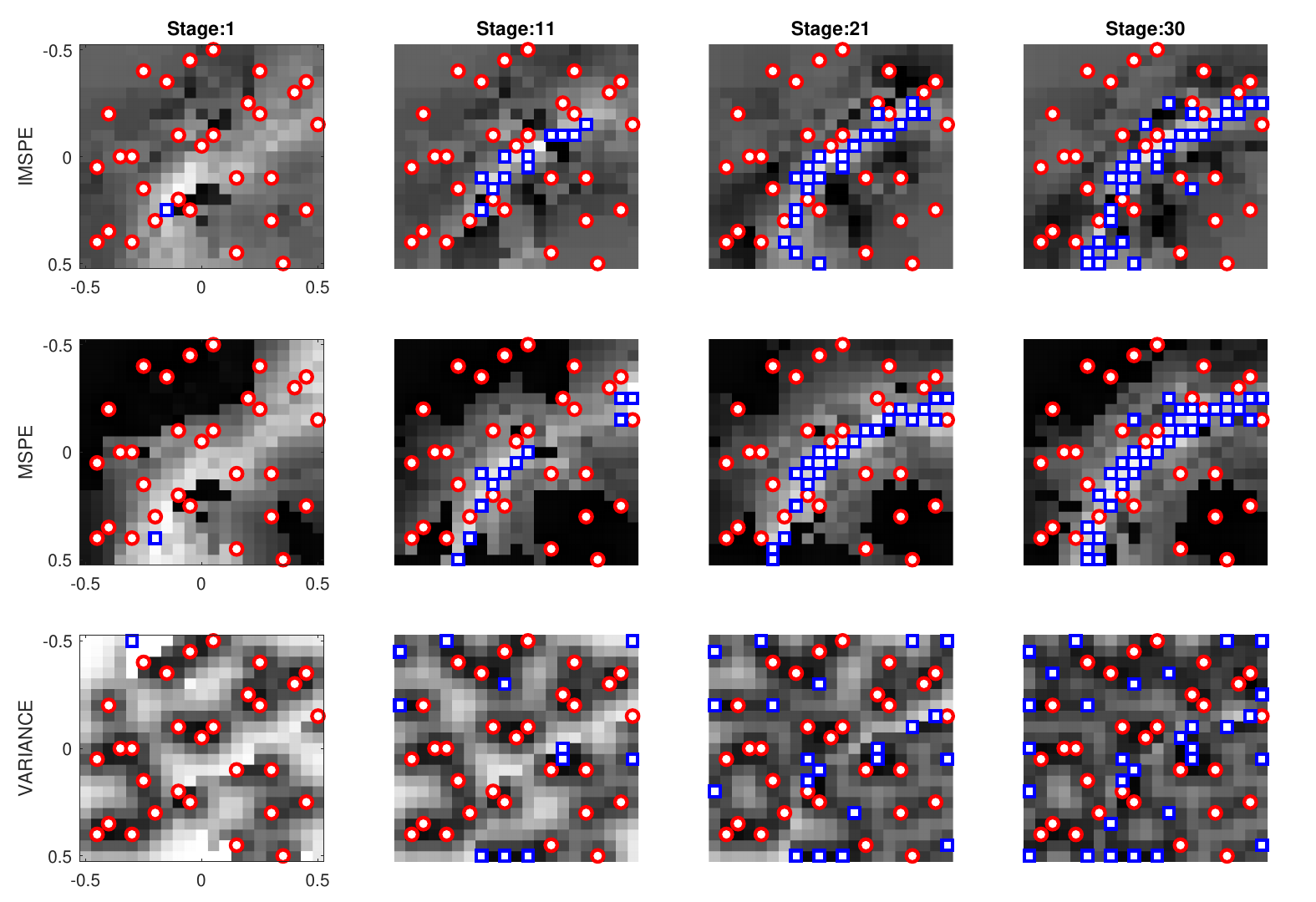}
	\vspace{-30pt}
	\caption{Active selection of design points for three acquisition functions. The background of each panel displays the acquisition function values, with lighter colors indicating higher values. Dots represent seed designs, and boxes represent the design points added by AL.} \label{fig13}
\end{figure}

For the same toy example, we ran AL for each choice of the three acquisition functions: start with a seed design of 30 random LHD points; subsequently add one additional design point every AL stage for 30 stages. Figure \ref{fig13} shows how the acquisition function values change as the AL stages progress and how they affect the selection training data inputs. For the IMSPE and MSPE criteria, the selected positions are highly concentrated around regional boundaries. With the variance criterion, the positions are close to a uniform distribution with a mild degree of concentration around the regional boundary. 

\section{Empirical benchmarking and validation} \label{sec:numerical}

In this section, we use simulation experiments and two real experiments to validate the proposed AL strategies for the JGP. Our metrics include out-of-sample mean root squared error (RMSE, smaller is better), the negative log posterior density score \citep[Eq.~\textbf{}25]{Gneiting2007} (NLPD, smaller is better), and the continuous ranked probability score \citep[Eq.~\textbf{}21]{Gneiting2007} (CRPS, smaller is better). We check how the three metrics change as more data are injected via AL. 

We compare our four criteria, JGP/MSPE, JGP/IMSPE, JGP/ALC and JGP/VAR, with the following benchmarks. First, a stationary GP with ALC, labeled as GP/ALC. We also considered an existing non-stationary GP model and its associated ALC criterion: Treed GP/ALC \citep{gramacy2008bayesian}  via the \texttt{tgp} package \citep{R-tgp}.  (We
omitted the two-layer Deep GP/ALC because it did not perform well in our small-$N$ setting.) For JGP, we set the local data size to $n=15$. Guidance on appropriate $n$
is provided by \citet[][Section 3.4]{park2022}. 

Each test problem is described in Section \ref{sec:datasets} with comparisons
between JGP methods in Section \ref{sec:comparison_1} and to non-JGP alternatives in \ref{sec:comparison_2}. We intentionally have the two separate comparison sections, because the RMSE values of JGP, GP and Treed GP are very different, and squeezing them into the same comparison plots does not show clear distinctions among closer performers. In Section \ref{sec:comparison_2}, we additional include JGP/MSPE as the best performer among the proposed criteria and JGP/ALC as contrast GP/ALC.

\subsection{Benchmark Datasets} \label{sec:datasets}

\paragraph{Bi-mixture GP datasets (BGP), $d = 2, 3, 4, 5$.}  We first work with synthetic examples in a rectangular domain $[-0.5, 0.5]^d$ of varying dimension.  The test functions are randomly sampled from a two-region partitioned GP model $f(\V{x}) = f_1(\V{x}) 1_{\mathcal{X}_1}(\V{x}) +  f_2(\V{x}) 1_{[0,1]^d \backslash \mathcal{X}_1}(\V{x}),$
where $\mathcal{X}_1 = \{ \V{a}^T\V{x} \ge 0\}$ with $\V{a}$ chosen uniformly at random in $\{-1, 1\}^d$, $f_1$ drawn from a zero-mean GP with variance $9$, and the isotropic Gaussian correlation function with length scale $0.1 \times d$, and $f_2$ from a GP with mean $13$, variance $9$, and correlation
function identical to $f_1$. Independent Gaussian noise with variance $4$ is added.

\begin{figure}[ht!]
	\centering
	\includegraphics[width=\textwidth]{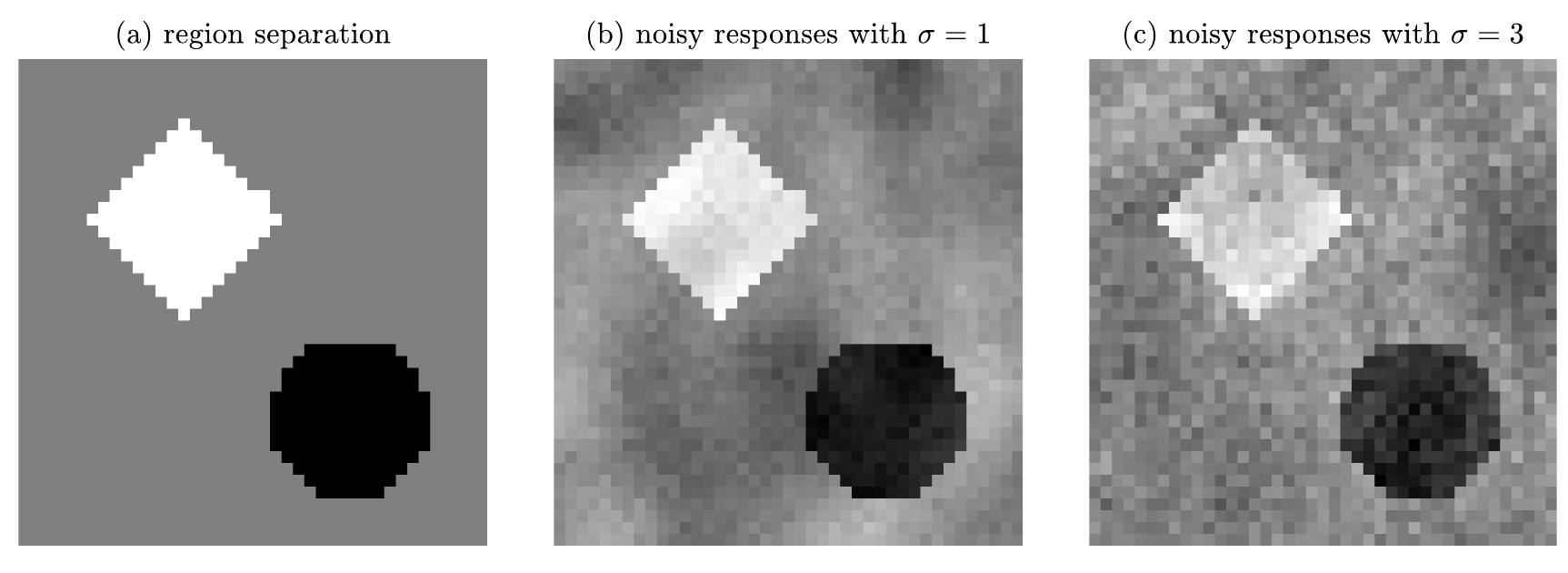} 
	\caption{MGP dataset. (a) region division with grayscale intensities proportional to prior GP means values, (b) and (c) the simulated noise responses with different noise levels 
		} \label{fig18}
\end{figure}

\paragraph{Multi-mixture GP datasets with complex boundaries (MGP), $d = 2$.} We also work with more complex synthetic examples in a rectangular domain $[-0.5, 0.5]^d$.  Due to the complexity of the regional boundaries and number of regions, we limit the dimension $d$ to two. For a larger $d$, one needs to run a large number of AL stages to explore the entire jump manifold. The test functions are randomly sampled from a three-region partitioned GP model with the domain partitioning depicted in Figure \ref{fig18}. Each of the three regions in the figure is independently sampled from a constant-mean GP with variance $9$, and the isotropic Gaussian correlation function with length scale $0.1 \times d$, and  constant mean values  proportional to the grayscale intensities appearing in Figure \ref{fig18} (a). Independent Gaussian noise with variance $\sigma^2$ is added,
and $\sigma^2 \in \{1^2, 3^2\}$ is varied to test the effect of the signal-to-noise ratio on AL performance. 

\paragraph{Smart factory dataset (SMF), $d=4$.} A semiconductor manufacturing facility employs an automated material handling system (AMHS) to optimize material flow between wafer fabrication steps. We perform a computer experiment to evaluate the AMHS under various design configurations \citep{kang2022}. A key performance metric associated with AMHS is the waiting time until an autonomous vehicle is finally assigned to serve a service request. We consider four design variables: vehicle acceleration, vehicle speed, the required minimum distance between vehicles, and a parameter determining vehicle dispatching policy. 
Preliminary experiments revealed that the waiting times are almost zero under most normal operating conditions, but can suddenly jump to significant levels as those conditions transition into ``heavy traffic'' situations. We posit that JGP with AL is a suitable approach to learn these jumping behaviors. We also observe from the preliminary study that the regime changes occurs along some axis-aligned directions. This regime change can be effectively modeled by both of JGP and TGP. 

\paragraph{Carbon nanotube yield dataset (CNT), $d=2$. } Here we present an application of the proposed JGP-AL strategy to a materials research problem, illustrating the applicability of the new approach. Due to high experimental expense, we can only illustrate how the new approach is applied to this real application.  Our colleagues at the Air Force Research Lab (AFRL) developed an autonomous research experimentation system (ARES) for carbon nanotubes \citep{Nikolaev2016}.  ARES is a robot capable of performing closed-loop iterative materials experimentation, carbon nanotube processing and ultimately measuring in-line process yield.   Here, we utilize  ARES to map out carbon nanotube yields as a function of two input conditions: reaction temperature and the log ratio of an oxidizing chemical concentration and a reducing chemical concentration. In previous experiments it was observed that nanotube yields are almost zero in certain growth conditions.  Then suddenly, when those conditions approach a regime where they become more activated, yield ``jumps'' up to a substantial level. These activating conditions vary significantly depending on how the two inputs we varied.

\subsection{Empirical evaluation} \label{sec:exp_config}
For the BGP and SMF, we set the seed design size to $N = 20 \times d$ and the number of the AL stages to 50. 
For CNT we were unable to conduct as many experiments as in other cases. The
to cope with slow runs, the experimentation system 
can perform multiple runs in parallel, making it amenable to AL in batches. We began with $N=20$ seed experiments, via LHD, and sequentially added batches of three new runs over five AL stages. 

A batch design of size $B$ with AL is defined as a collection of $B$ future design points $\{\V{x}_{N+1}, \dots, \V{x}_{N+B}\}$ selected by a chosen AL criterion, based on past $N$ experimental observations $\{(\V{x}_{i}, y_i), i = 1,\dots, N\}$, without collecting any additional experimental results. The selection of each design point in the batch is achieved iteratively. The selection of the $(k+1)^\mathrm{st}$ design point $\V{x}_{N+k+1}$ in the batch is based on the AL criterion evaluated with all past experimental observations $\{(\V{x}_{i}, y_i), i = 1,\dots, N\}$, plus imputed observations at the previously selected design points in the same batch $\{(\V{x}_{N+j}, \hat{y}_{N+j}), j = 1,\dots,k\}$, where $\hat{y}_{N+j}$ is the predictive mean of JGP evaluated with the past $N$ observations. This imputation prevents the design points in the same batch from being too closely located.

\label{sec:comparison_1}

\begin{figure}[ht!]
	\centering
	\includegraphics[width=\textwidth]{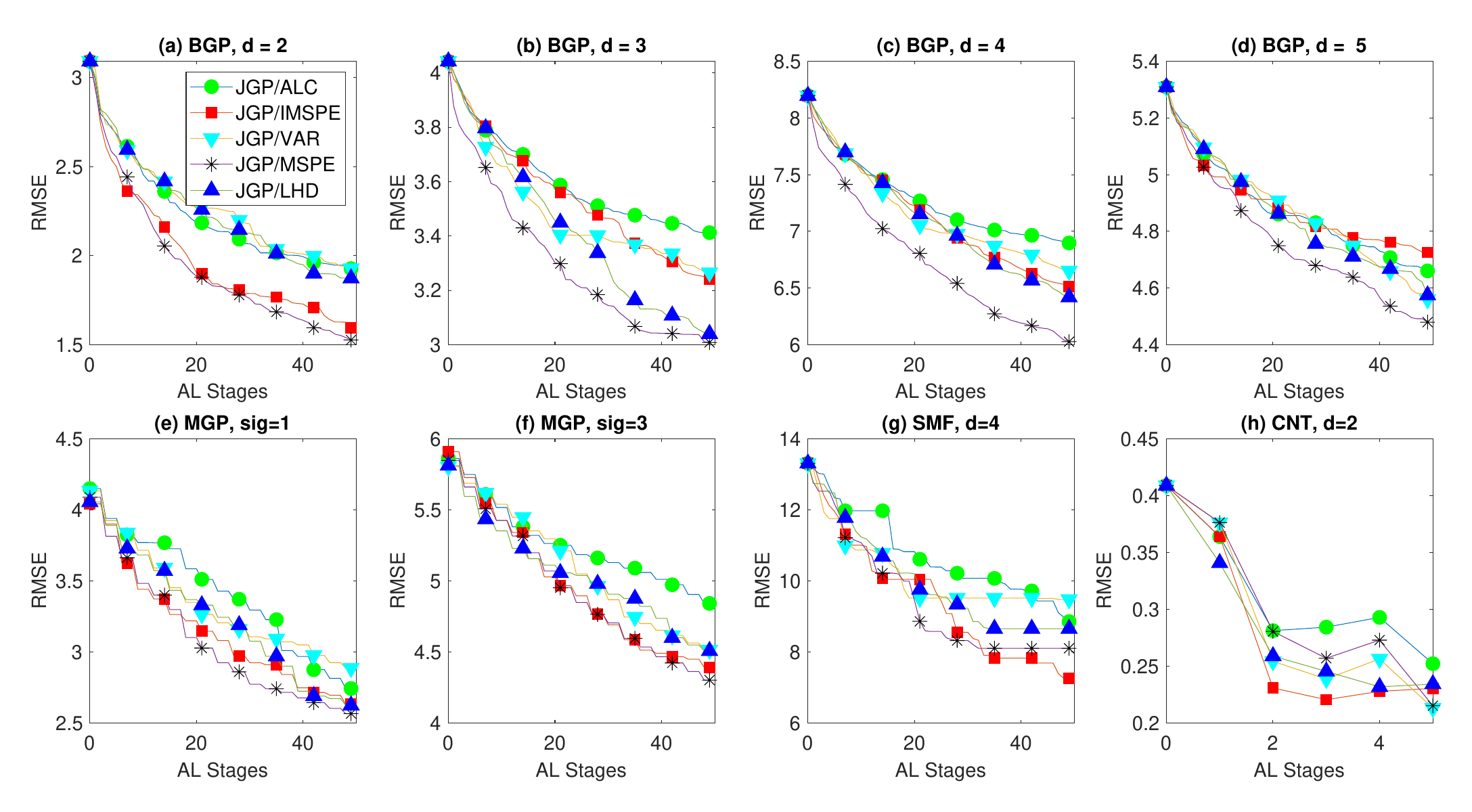}
	\vspace{-30pt}
	\caption{RMSE of JGP/ALC, JGP/IMSPE, JGP/VAR, JGP/MSPE, and JGP/LHD. Panels (a--d) values reported are based on 50 MC repetitions. Panel (f) is based on a batch AL with three experiments per each AL stage.} \label{fig_new_1}
\end{figure}

Figures \ref{fig_new_1} depicts the trend of RMSE values as AL stages progress. For the simulated BGP and MGP datasets, we average the values over 50 Monte Carlo (MC) replications, while for the other real datasets, the values are based on one experimental campaign. The comparison reveals that incorporating bias into the design of AL criteria significantly enhances the accuracy of JGP. Specifically, For all datasets except CNT, JGP/MSPE—which includes a bias term in addition to the variance considered by JGP/VAR—outperforms JGP/VAR. Paired Wilcoxon tests confirm the statistical significance of the results [see Appendix D].

For CNT, JGP/MSPE and JGP/VAR exhibit similar performance with comparable choices of design points. To further investigate, we examined the bias and variance estimates of JGP for this dataset. We discovered that the variance values are significantly larger than the squared bias values. Consequently, both JGP/MSPE and JGP/VAR have a similar ability reduce variance with acquisitions. This does not imply that both lead to a space-filling design, unlike the variance-based AL criterion of the stationary GP. As we illustrated in Section \ref{sec:AL_illus}, JGP/VAR selects distinct designs compared to GP/ALC, as it tends to concentrate more around the regional boundaries. See Figure \ref{fig_new_3}. 

\begin{figure}[ht!]
	\centering
	\includegraphics[width=\textwidth]{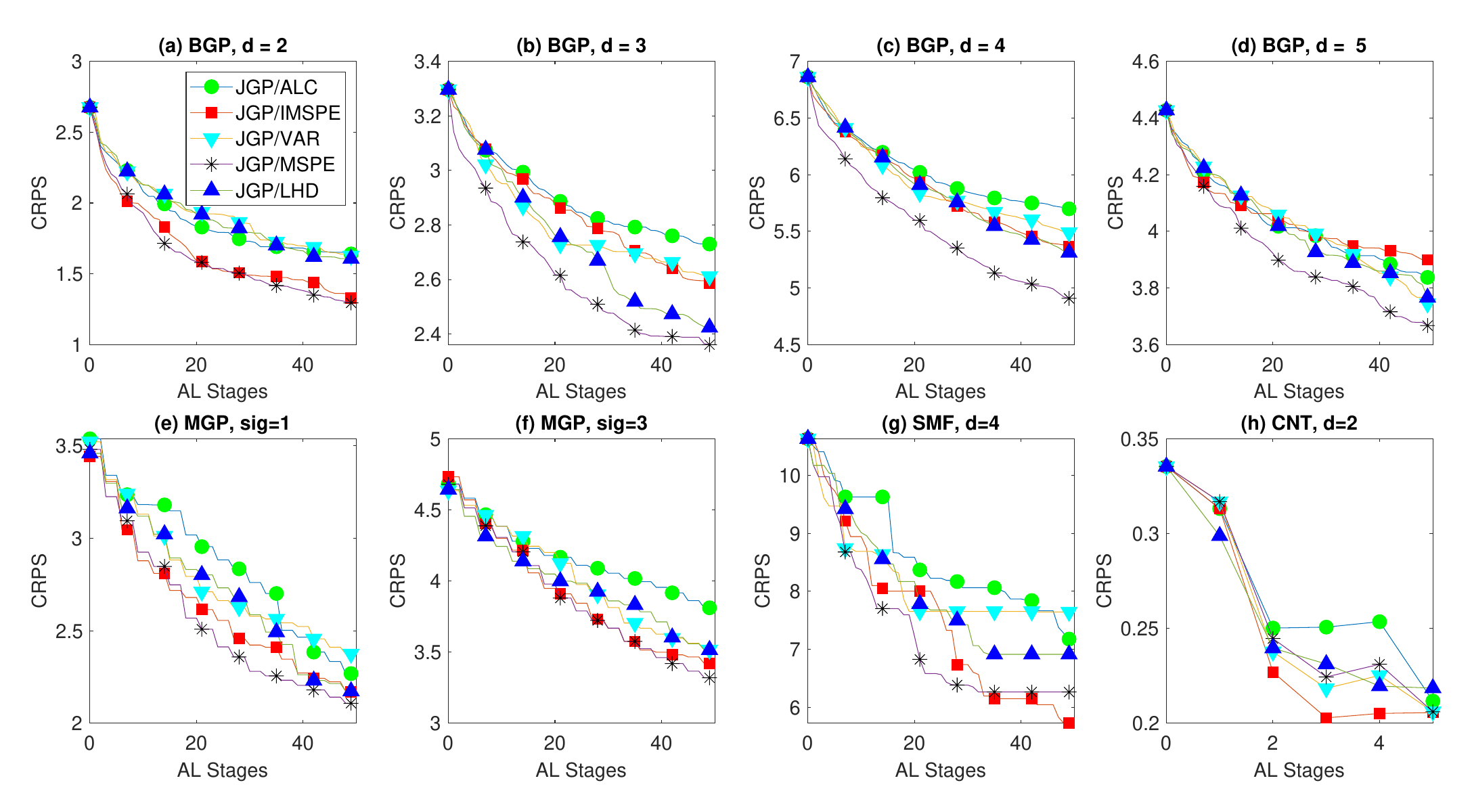}
	\vspace{-30pt}
	\caption{CRPS of JGP/ALC, JGP/IMSPE, JGP/VAR, JGP/MSPE, and JGP/LHD. 
	} \label{fig_crps1}
\end{figure}

\begin{figure}[ht!]
	\centering
	\includegraphics[width=\textwidth]{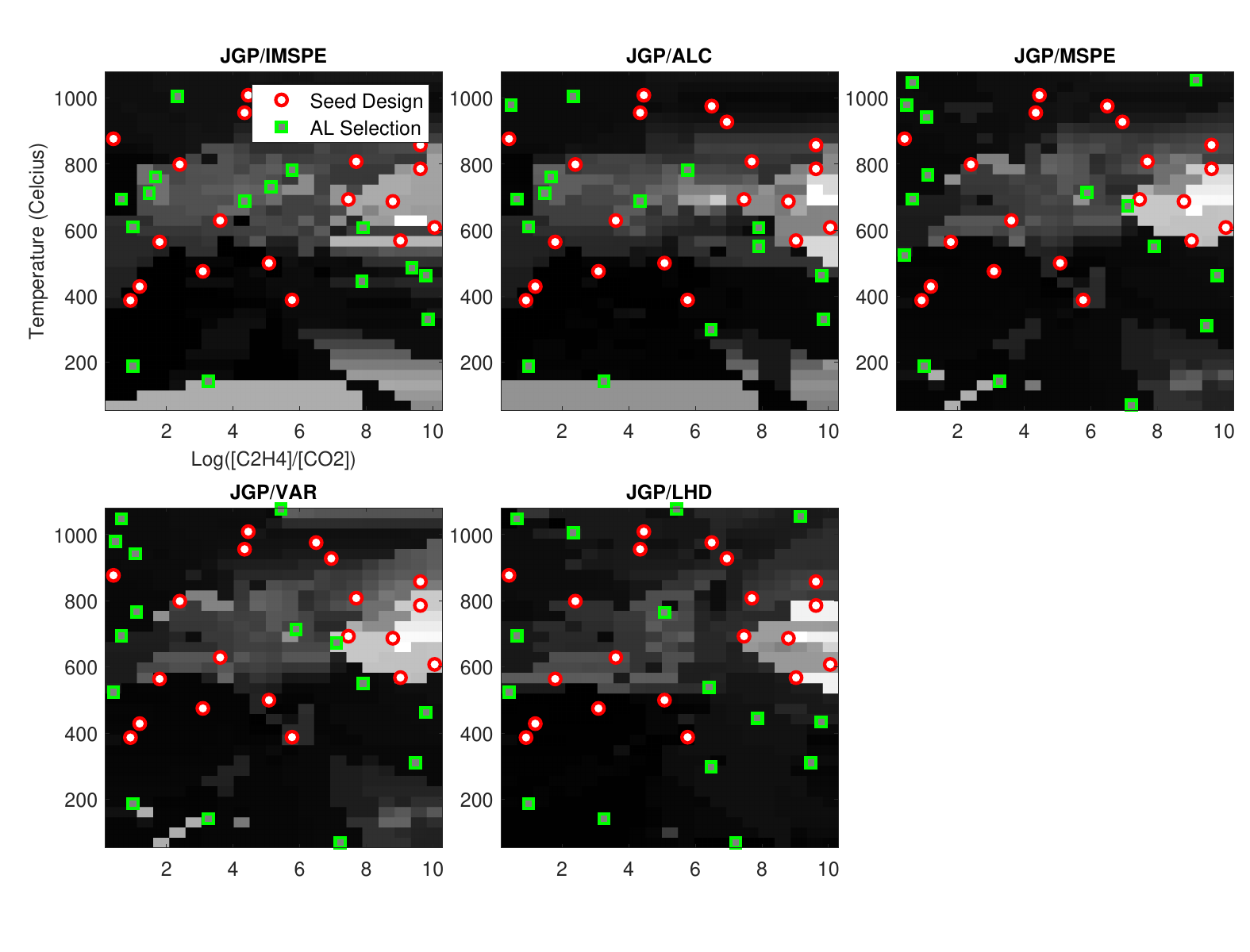}
	\vspace{-30pt}
	\caption{AL of JGP for CNT with the proposed criteria. In all panels, the background displays the acquisition function values, with lighter colors indicating higher values.} \label{fig_new_3}
\end{figure}

Another intriguing comparison involves JGP/MSPE, JGP/IMSPE, and JGP/LHD. The first two are based on AL strategies, while the last relies on a simple space-filling design. In simulated cases, JGP/MSPE outperform JGP/LHD with statistically significant margins for all of the datasets. Additionally, JGP/IMSPE and JGP/MSPE outperform JGP/LHD for the SMF dataset. Notably, for the CNT dataset, JGP/IMSPE achieves a faster RMSE convergence to the bottom level compared to JGP/LHD.

\begin{figure}[ht!]
	\centering
	\includegraphics[width=\textwidth]{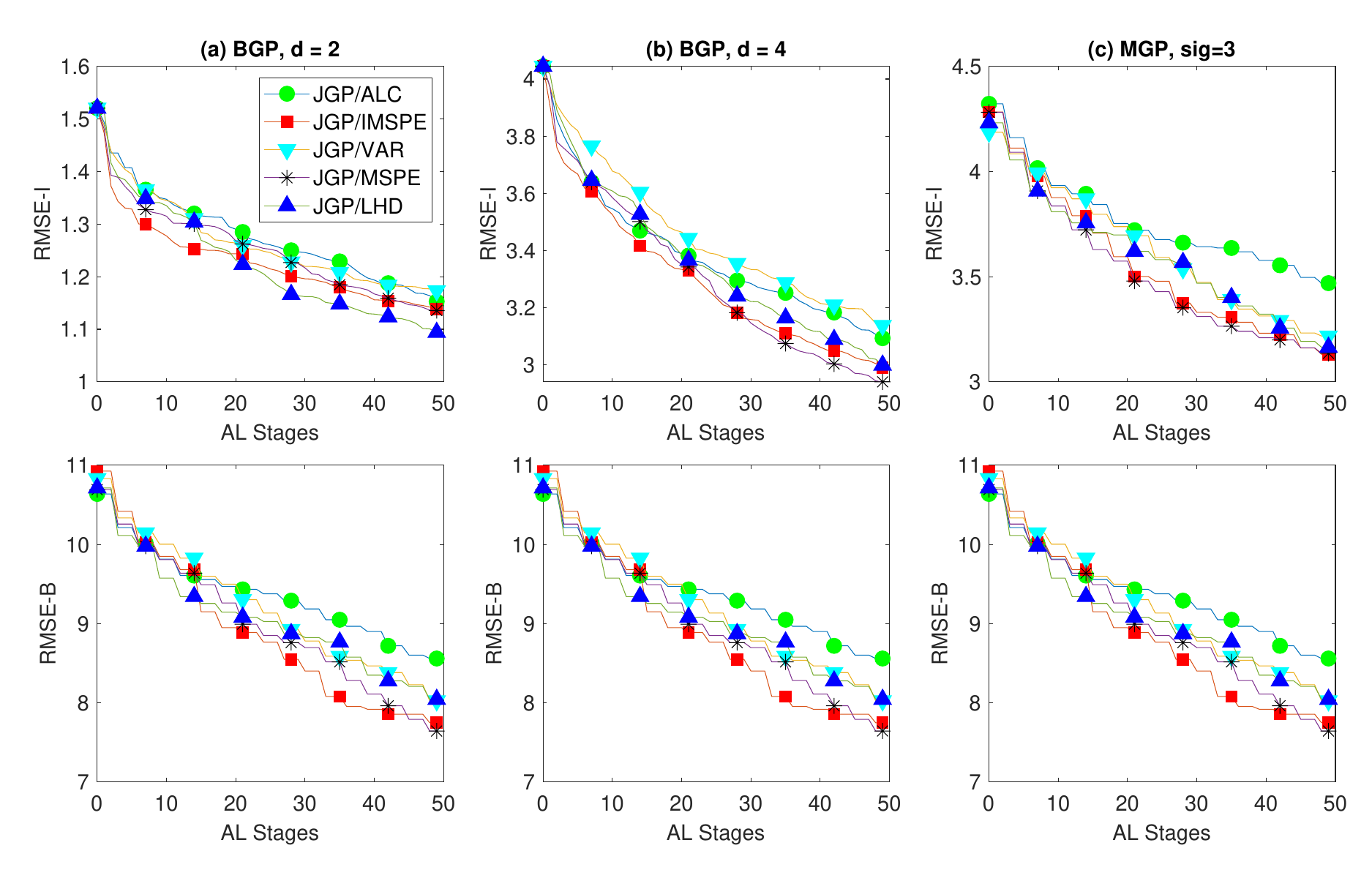}
	\vspace{-30pt}
	\caption{Breakdown of the RMSE values into RMSE-B (RMSE on/near boundaries) and RMSE-I (RMSE on interior areas)} \label{fig_RMSE_Bnd}
\end{figure}

We also investigated the sources of RMSE improvements for JGP/MSPE and JGP/IMSPE by decomposing the overall RMSE into two components: RMSE near or on the boundaries (RMSE-B) and RMSE in the interior regions (RMSE-I). Figure \ref{fig_RMSE_Bnd} illustrates the trends of these two components for three of the simulated datasets. We observe that RMSE-I values are similar across all compared AL strategies, while JGP/MSPE and JGP/IMSPE exhibit notably lower RMSE-B values. This suggests that the majority of the RMSE improvement can be attributed to enhanced prediction accuracy near the boundaries. Similar trends are observed in the other simulated datasets.

To assess UQ capabilities we use two widely adopted metrics: Negative Log Posterior Density (NLPD) and Continuous Ranked Probability Score (CRPS). Both evaluate the quality of probabilistic predictions by measuring how well the Gaussian predictive distribution fits the test dataset. For each test observation $(\V{x}_*, y_*)$, the CRPS for the Gaussian predictive distribution  $\mathcal{N}(\hat{\mu}, \hat{s}^2)$ is evaluated as
\begin{equation*}
	CRPS(\hat{\mu}, \hat{s}^2,  y_*) = \sigma \left[\tilde{y}_{*} \left(2\Phi\left(\tilde{y}_{te}\right)-1\right)+2 \phi\left(\tilde{y}_{te}\right) -\frac{1}{\sqrt{\pi}}  \right],
\end{equation*}
where $\phi$ and $\Phi$ denote the probability density function and cumulative distribution function of a standard Gaussian random variable, and $\tilde{y}_{*} = (y_{*}-\hat{\mu})/\hat{s}$. The NLPD is defined as 
\begin{equation*}
	NLPD(\hat{\mu}, \hat{s}^2,  y_{*}) = -\log \left[\frac{1}{\hat{s}} \phi\left(\tilde{y}_{*}\right) \right].
\end{equation*}
The reported NLPD and CRPS scores are averaged over all test locations. Figure \ref{fig_crps1} show the trend of the CRPS scores as the AL progresses, which are consistent to those of RMSE. The NLPD results showed similar trend. Due to the page limitation, we place it in Appendix F. Figure 4. 

\begin{figure}[ht!]
	\centering
	\includegraphics[width=\textwidth]{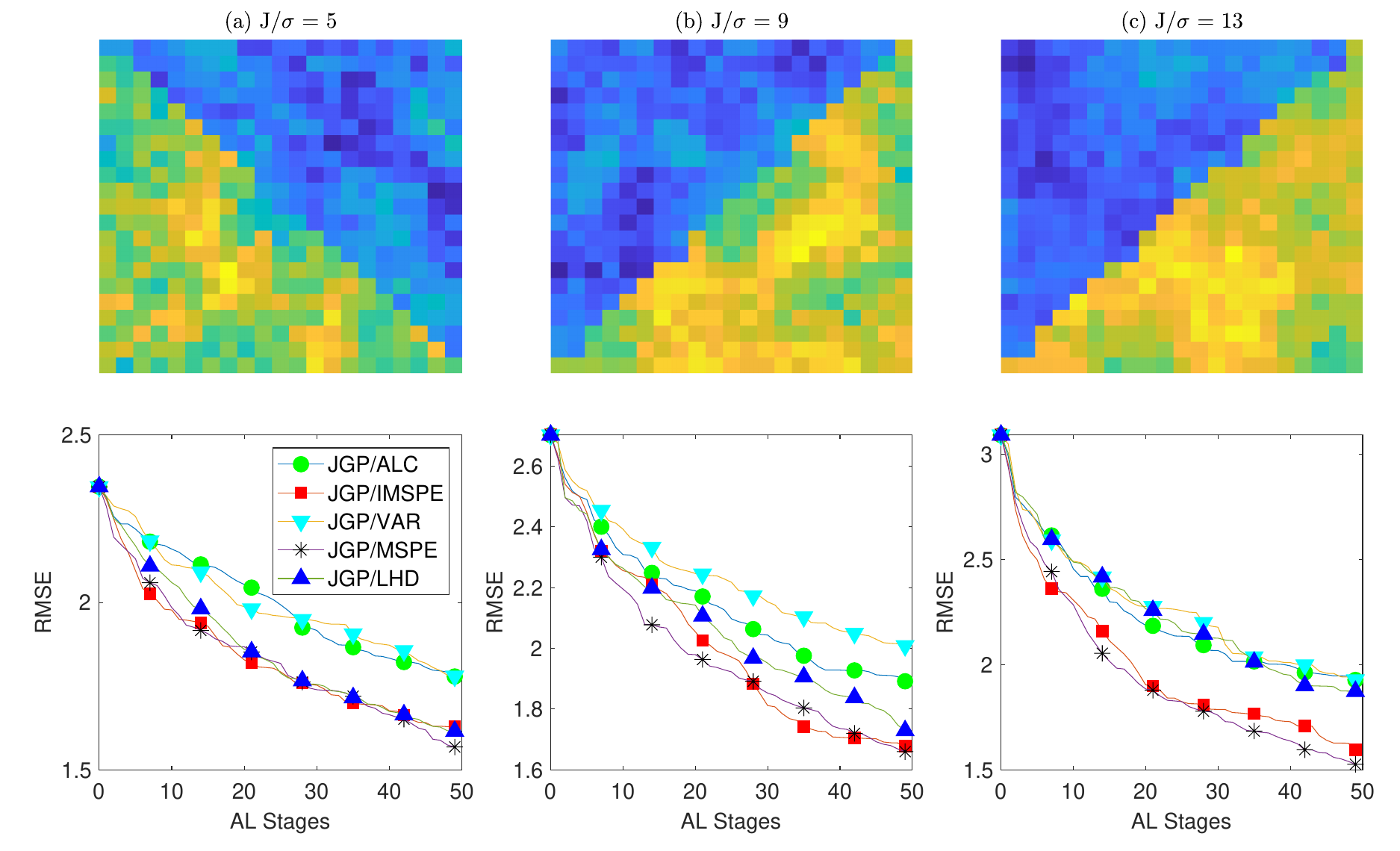}
	\vspace{-30pt}
	\caption{RMSEs for different jump sizes $J$ relative to the noise level $\sigma$.} \label{fig_RMSE_J}
\end{figure}

We also evaluated how the different AL strategies perform across varying jump sizes $J$. Specifically, we used the BGP dataset with $d=2$, varying the jump size relative to the noise standard deviation ($J/\sigma$) from 5 to 13. As shown in the top panels of Figure \ref{fig_RMSE_J}, the jump in the response surface is not clearly visible at $J/\sigma=5$ due to the noise. The bottom panels illustrate RMSE trends under different $J/\sigma$ conditions. For the smallest jump size, JGP/MSPE and JGP/IMSPE perform similarly to the space-filling design, JGP/LHD, as the bias in the design choices has limited influence. However, as the jump size increases, JGP/MSPE and JGP/IMSPE increasingly outperform the space-filling design and other variance-based criteria. These results suggest that incorporating bias into the AL strategy becomes more advantageous in scenarios with larger jumps.

Next we considered MGP with different noise levels: $\sigma^2 = 1^2$ (low noise) and $\sigma^2 = 3^2$ (high noise). As illustrated in Figure \ref{fig18}, the dataset with $\sigma^2 = 1^2$ represents a low-noise scenario, while the dataset with $\sigma^2 = 3^2$ represents a high-noise scenario. Yet Figures \ref{fig_new_1}, and \ref{fig_crps1} show similar trends in the metrics. 

\subsection{Comparison to Treed GP and Stationary GP} \label{sec:comparison_2}
We compare the best performers among the proposed AL criteria — JGP/MSPE and JGP/IMSPE — with TGP/ALC and GP/ALC. Note that our comparators not only differ in their AL criteria, but also in their modeling approaches.  So differences in RMSE, NLPD and CRPS are not purely due to the choices of training data. Throughout this comparison, we aim to demonstrate that JGP with AL that incorporates bias is better at learning piecewise continuous surrogates than other non-stationary and stationary models. 

\begin{figure}[ht!]
	\centering
	\includegraphics[width=\textwidth]{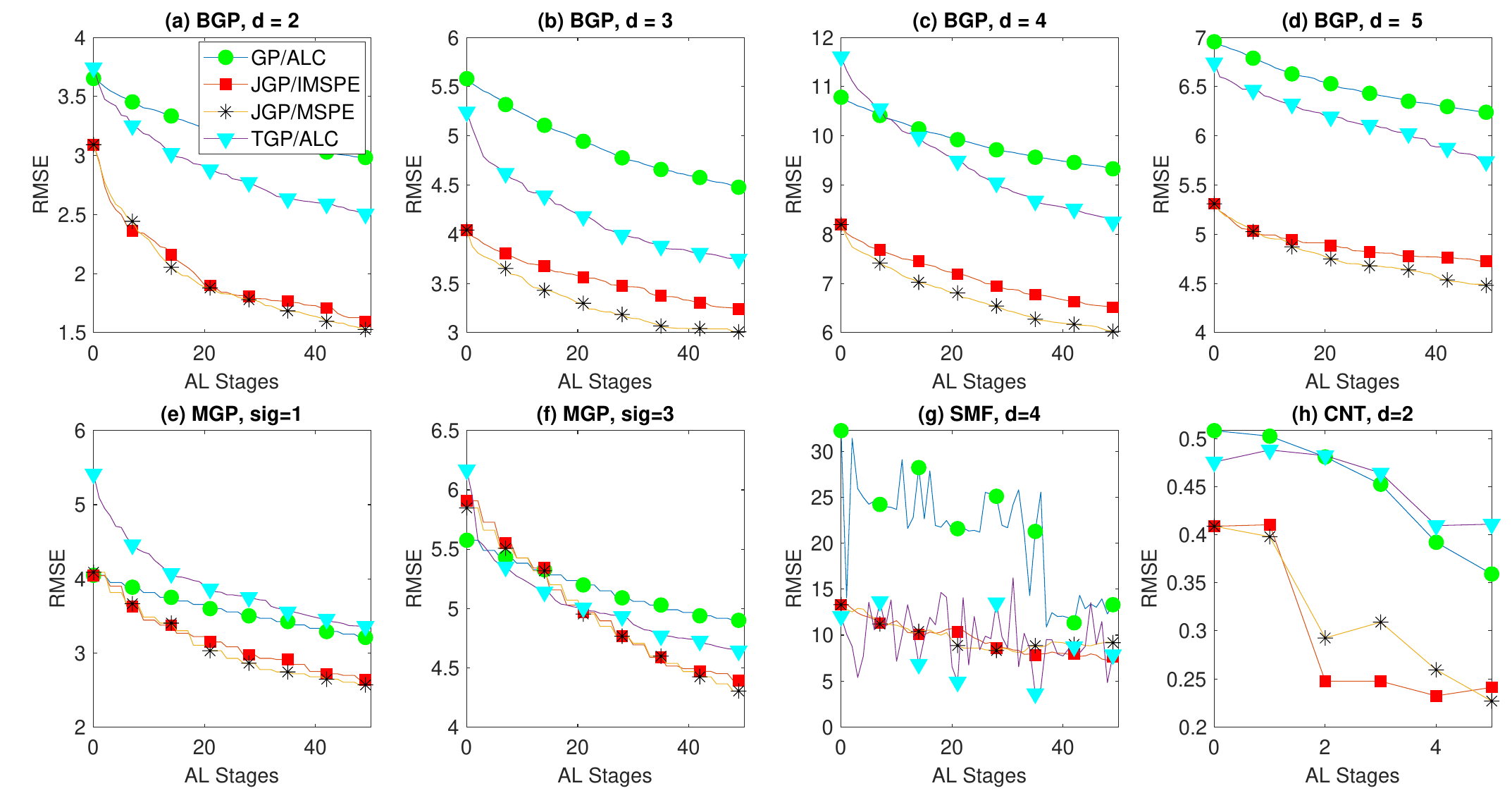}
	\vspace{-30pt}
	\caption{RMSE of JGP/IMSPE, JGP/MSPE, TGP/ALC and GP/ALC. 
	} \label{fig_new_4}
\end{figure}

\begin{figure}[ht!]
	\centering
	\includegraphics[width=\textwidth]{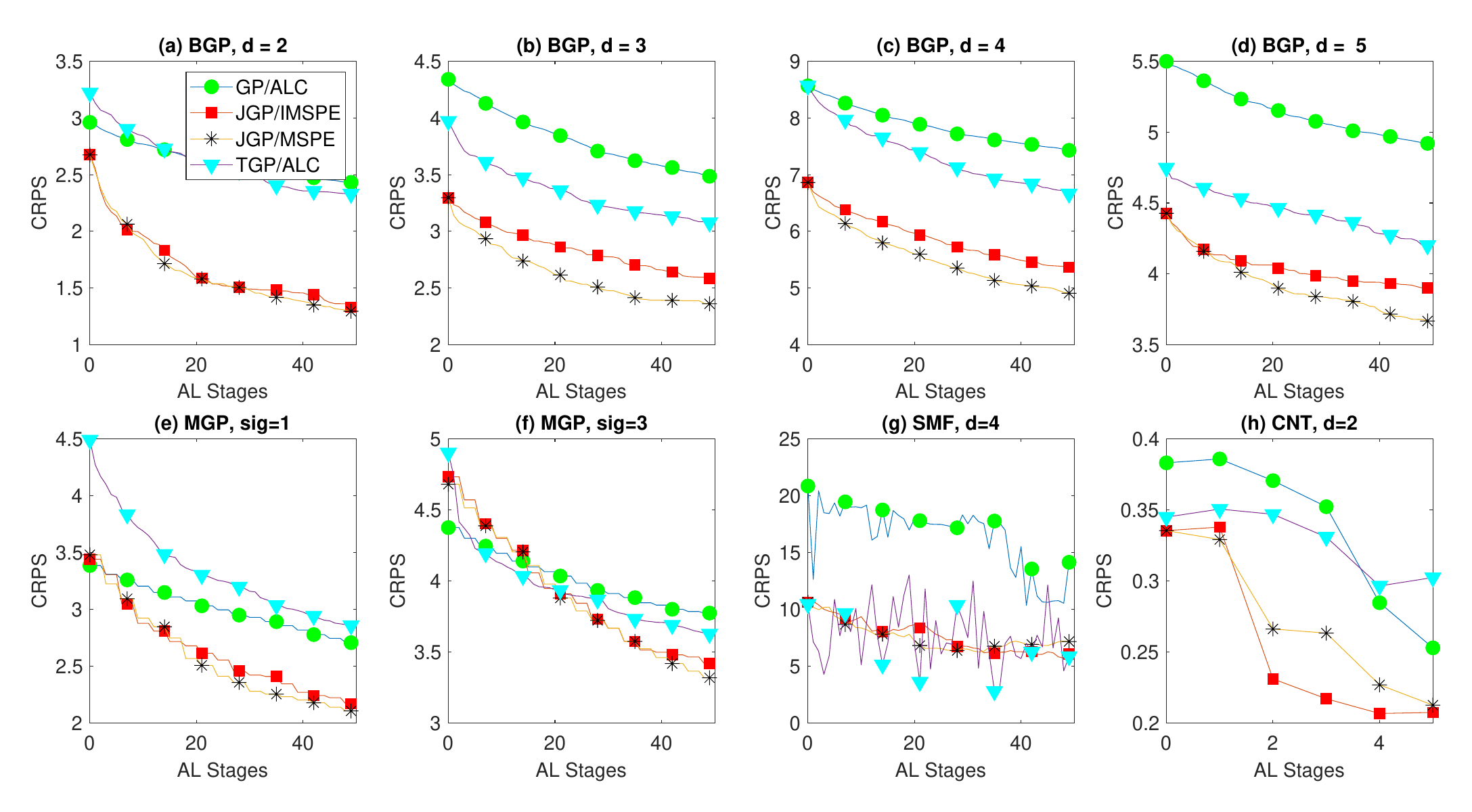}
	\vspace{-30pt}
	\caption{CRPS of JGP/IMSPE, JGP/MSPE, TGP/ALC and GP/ALC. 
	} \label{fig_new_7}
\end{figure}

Figures \ref{fig_new_4}, and \ref{fig_new_7} show the trends of RMSE and CRPS scores as AL progresses for eight test cases. Observe that the two non-stationary models outperform GP/ALC significantly by all three metrics and all eight test cases. Pairwise Wilcoxon tests reveal statsitical significance [Appendix E]. This demonstrates the advantage of using nonstationary GP models versus the conventional stationary GP model for data from these particular processes.

\begin{figure}[ht!]
	\centering
	\includegraphics[width=0.6\textwidth]{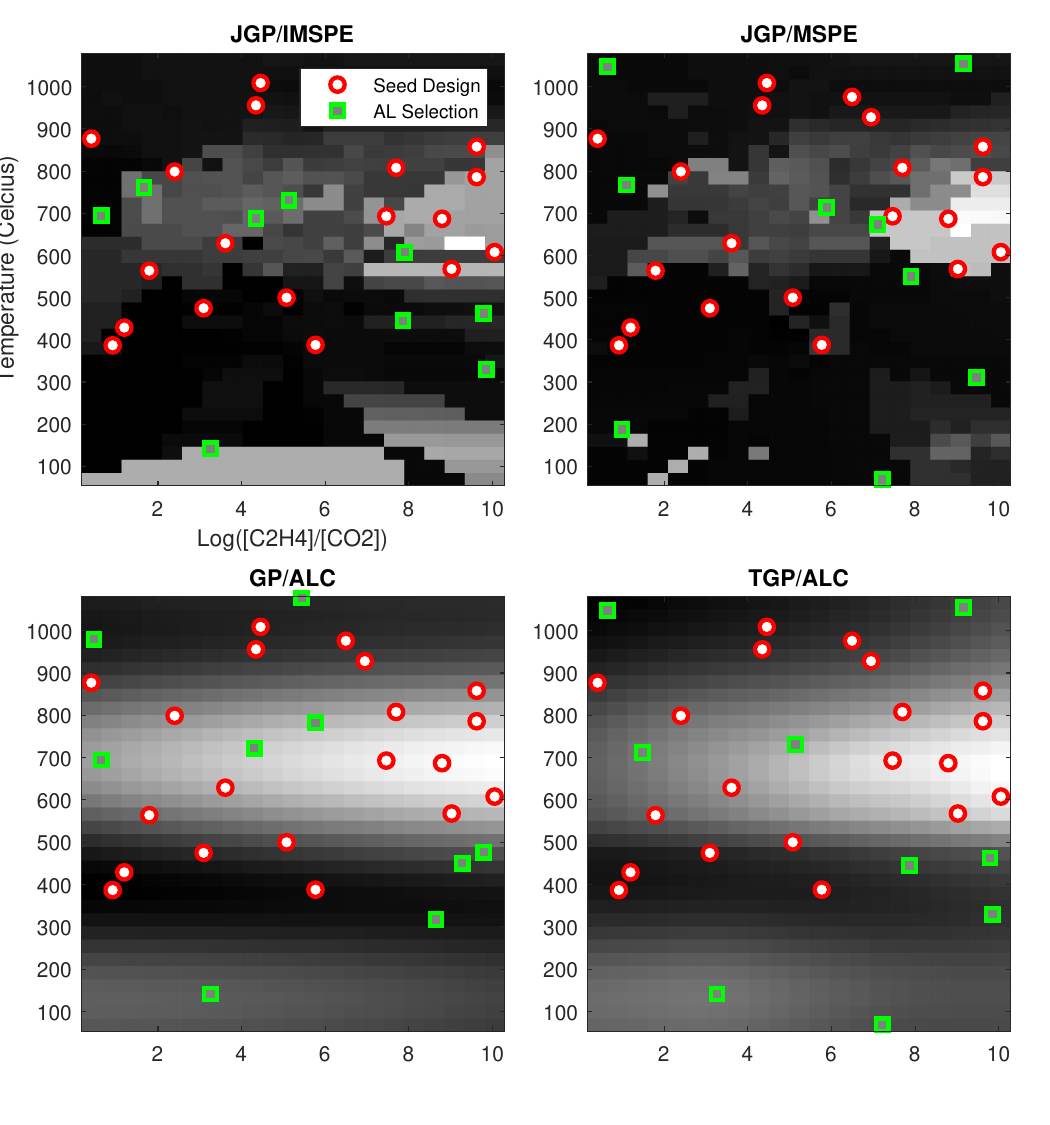}
	\vspace{-30pt}
	\caption{AL of JGP, GP and TGP for CNT dataset after the third AL stage. In all  panels the background displays the acquisition function with lighter colors indicating higher values.} \label{fig_new_6}
\end{figure}

It is also interesting to note that JGP/MSPE and JGP/IMSPE outperform TGP/ALC with significant margins in RMSE and CRPS for the BGP, MGP and CNT datasets. For those, TGP/ALC performs more closely to GP/ALC. This can be understood through Figure \ref{fig_new_6}, which shows the AL-based design selections for CNT. 
TGP/ALC selects design points similarl to GP/ALC. For SMF, JGP/MSPE and JGP/IMSPE perform comparably to TGP/ALC. As we noted in the data description, the SMF process undergoes regime changes along coordinate-wise directions, which is ideal for TGP.


We also like to note that the NLPD [Appendix F. Figure 5] and CRPS scores [Figure \ref{fig_new_7}] exhibit inverse trends. While JGP/MSPE and JGP/IMSPE achieve significantly better overall CRPS scores than TGP/ALC, the NLPD scores for TGP/ALC are substantially lower than those of JGP/MSPE and JGP/IMSPE. These low NLPD values for TGP/ALC are primarily due to its high predictive variance estimates. For instance, on the BGP dataset with $d=5$, TGP/ALC attains much lower NLPD scores, with a median variance estimate of approximately 100, compared to about 12 for JGP/MSPE. These inflated variance estimates shadow the high RMSE issues with TGP/ALC, rendering the NLPD metric potentially misleading.

\section{Conclusion} \label{sec:conc}
We explored the Jump GP model for piecewise continuous response surfaces, focusing on predictive bias and variance estimates to develop an effective active learning (AL) heuristic. We showed that JGP bias is largely influenced by the accuracy of the local classifiers estimating regime changes, whereas model variance is comparable to that of the standard GP model. To reduce model bias and variance together we should invest more data points (via AL) around boundaries between regimes, while continuing to place data points around less populated areas of a design space. Based on that principle, we introduced four AL criteria: JGP/IMSPE (minimizing integrated mean squared prediction error), JGP/ALC (minimizing integrated predictive variances), JGP/MSPE (placing points at the peak of mean squared prediction error), and JGP/VAR (placing at the peak of predictive variance).

We evaluated these four comparators using various simulation scenarios by tracking the changes in root mean square prediction  error (RMSE), negative log posterior density (NLPD),  continuous ranked probability score (CRPS) and other metrics. JGP/MSPE and JGP/IMSPE outperformed the variance-based criteria (JGP/ALC and JGP/VAR). Comparing JGP/MSPE and JGP/IMSPE to the stationary GP with ALC and a non-stationary Bayesian treed GP (TGP) with ALC, both JGP and TGP outperformed GP, highlighting the need for non-stationary models. JGP outperformed TGP for most of the test cases. 

Finally, we illustrated the applicability of our methods with two real experiments. Our first example involved a study of the performance of an autonomous material handling system in smart factory facilities. In the second experiment we highlighted a real materials design application of the proposed method for effectively mapping carbon nanotube yield as a function of two design variables. In this example, JGP/MSPE and JGP/IMSPE outperformed TGP and GP with a significant margin.

\section*{Supplementary Material}
\begin{description}
	
	\item[Online Supplemental Material:] A single PDF document provides supplementary material, including the proof of Theorem 1 [Appendix A]; an empirical validation of the bias and variance estimates presented in Section 3 [Appendix B]; an empirical validation on the distribution of $\hat{p}_i$ [Appendix C]; paired Wilcoxon tests assessing the statistical significance of performance comparisons among JGP/ALC, JGP/IMSPE, JGP/VAR, JGP/MSPE, and JGP/LHD [Appendix D]; paired Wilcoxon tests for performance comparisons among JGP/IMSPE, JGP/MSPE, TGP/ALC, and GP/ALC [Appendix E]; and the NLPD scores for all numerical cases [Appendix G].
	
	\item[MATLAB package for ''Active Learning for Jump GP":] MATLAB-package containing code to perform the active learning method described in the article. The package also contains simulation datasets used as examples in the article. (a zip file)
\end{description}

\section*{Data Availability Statement}
To support reproducibility, the simulation data generation codes used in this study are shared along with the source code. Access to the real-world datasets can be requested from the lead author and will be granted pending necessary security clearances.

\section*{Conflict of Interest Statement} 
The authors report there are no competing interests to declare. 

\if0\blind
{
\section*{Acknowledgment} 
We acknowledge support for this work. Park and Gramacy are supported by the National Science Foundation (NSF-2420358, NSF-2152679). Waelder and Maruyama are supported by the Air Force Office of Scientific Research (LRIR-19RXCOR040). Kang and Hong are partially supported by the National Research Foundation of Korea (NRF-2020R1A2C2004320) and by the BK21 FOUR of the National Research Foundation of Korea (NRF-5199990914451). The main algorithm of this work is protected by a patent pending with application number 18/532,296.
} \fi

%
%
%
\spacingset{1.0}
\bibliographystyle{agsm}
\bibliography{jumpgp}

@Article{beck2016sequential,
	author    = {Beck, Joakim and Guillas, Serge},
	journal   = {SIAM/ASA Journal on Uncertainty Quantification},
	title     = {Sequential design with mutual information for computer experiments ({MICE}): Emulation of a tsunami model},
	year      = {2016},
	number    = {1},
	pages     = {739--766},
	volume    = {4},
	publisher = {SIAM},
}

@Manual{R-tgp,
	title = {{\tt tgp}: {B}ayesian Treed {G}aussian Process Models},
	author = {Gramacy, RB and Taddy, MA},
	year = {2016},
	note = {R package version 2.4-14}
}

@book{wendland2004scattered,
	title={Scattered Data Approximation},
	author={Wendland, H},
	year={2004},
	publisher={Cambridge University Press},
	address={Cambridge, England}
}

@book{santner2018design,
	Author = {Santner, TJ and Williams, BJ and Notz, W},
	Title = {The Design and Analysis of Computer Experiments, Second Edition},
	Year = {2018},
	Publisher = {Springer--Verlag},
	Pages = {436},
	address = {New York, NY}
}

@article{lin2015latin,
	title={Latin hypercubes and space-filling designs},
	author={Lin, CD and Tang, B},
	journal={Handbook of Design and Analysis of Experiments},
	pages={593--625},
	year={2015},
	publisher={NewYork: Chapman and Hall/CRC}
}

@book{williams2006gaussian,
	title={Gaussian Processes for Machine Learning},
	author={Rasmussen, CE and Williams, CKI},
	year={2006},
	publisher={MIT Press},
	address={Cambridge, MA}
}

@Article{bryant1978asymptotic,
	author    = {Bryant, Peter and Williamson, John A},
	journal   = {Biometrika},
	title     = {Asymptotic behaviour of classification maximum likelihood estimates},
	year      = {1978},
	number    = {2},
	pages     = {273--281},
	volume    = {65},
	publisher = {Oxford University Press},
}

@Article{cohn1996active,
	author  = {Cohn, David A and Ghahramani, Zoubin and Jordan, Michael I},
	journal = {Journal of Artificial Intelligence Research},
	title   = {Active learning with statistical models},
	year    = {1996},
	pages   = {129--145},
	volume  = {4},
	number  = {1}
}

@InProceedings{damianou2013deep,
	title = 	 {Deep {G}aussian Processes},
	author = 	 {Damianou, Andreas and Lawrence, Neil D.},
	booktitle = 	 {Proceedings of the Sixteenth International Conference on Artificial Intelligence and Statistics},
	pages = 	 {207--215},
	year = 	 {2013},
	editor = 	 {Carvalho, Carlos M. and Ravikumar, Pradeep},
	volume = 	 {31},
	series = 	 {Proceedings of Machine Learning Research},
	address = 	 {Scottsdale, Arizona, USA},
	publisher =    {PMLR}
}

@Article{Gneiting2007,
	author    = {Tilmann Gneiting and Adrian E Raftery},
	journal   = {Journal of the American Statistical Association},
	title     = {Strictly Proper Scoring Rules, Prediction, and Estimation},
	year      = {2007},
	number    = {477},
	pages     = {359-378},
	volume    = {102},
	publisher = {Taylor & Francis}
}

@Book{gramacy2020surrogates,
	author    = {Gramacy, Robert B},
	publisher = {CRC Press},
	title     = {Surrogates: {Gaussian} Process Modeling, Design, and Optimization for the Applied Sciences},
	year      = {2020},
	address   = {Boca Raton, FL, USA}
}

@Article{gramacy2015local,
	author    = {Gramacy, Robert B and Apley, Daniel W},
	journal   = {Journal of Computational and Graphical Statistics},
	title     = {Local {Gaussian} process approximation for large computer experiments},
	year      = {2015},
	number    = {2},
	pages     = {561--578},
	volume    = {24},
	publisher = {Taylor \& Francis},
}

@Article{gramacy2009adaptive,
	author    = {Gramacy, Robert B and Lee, Herbert KH},
	journal   = {Technometrics},
	title     = {Adaptive design and analysis of supercomputer experiments},
	year      = {2009},
	number    = {2},
	pages     = {130--145},
	volume    = {51},
	publisher = {Taylor \& Francis},
}

@article{gramacy2014massively,
	title={Massively parallel approximate {G}aussian process regression},
	author={Gramacy, RB and Niemi, J and Weiss, RM},
	journal={SIAM/ASA Journal on Uncertainty Quantification},
	volume={2},
	number={1},
	pages={564--584},
	year={2014},
	publisher={SIAM}
}

@article{binois2018practical,
	author    = {Binois, M and Gramacy, RB and Ludkovski, M},
	title     = {Practical heteroscedastic {G}aussian process modeling for large simulation experiments},
	journal   = {Journal of Computational and Graphical Statistics},
	year      = {2018},
	volume    = {27},
	number    = {4},
	pages     = {808--821},
	doi       = {10.1080/10618600.2018.1458625},
	eprint    = {https://doi.org/10.1080/10618600.2018.1458625},
	publisher = {Taylor \& Francis},
}

@Article{gramacy2008bayesian,
	author    = {Gramacy, Robert B and Lee, Herbert K H},
	journal   = {Journal of the American Statistical Association},
	title     = {Bayesian treed {Gaussian} process models with an application to computer modeling},
	year      = {2008},
	number    = {483},
	pages     = {1119--1130},
	volume    = {103},
	publisher = {Taylor \& Francis},
}

@Article{heaton2017nonstationary,
	author    = {Heaton, Matthew J and Christensen, William F and Terres, Maria A},
	journal   = {Technometrics},
	title     = {Nonstationary {Gaussian} process models using spatial hierarchical clustering from finite differences},
	year      = {2017},
	number    = {1},
	pages     = {93--101},
	volume    = {59},
	publisher = {Taylor \& Francis},
}

@Article{kang2022,
	author  = {Kang, Bonggwon and Park, Chiwoo and Kim, Haejoong and Hong, Soondo},
	journal = {IEEE Transactions on Automation Science and Engineering},
	title   = {Bayesian Optimization for the Vehicle Dwelling Policy in a Semiconductor Wafer Fab},
	year    = {2024},
	pages   = {5942--5952},
	volume  = {21},
	number  = {4}
}

@Article{kim2005analyzing,
	author    = {Kim, Hyoung-Moon and Mallick, Bani K and Holmes, Chris C},
	journal   = {Journal of the American Statistical Association},
	title     = {Analyzing nonstationary spatial data using piecewise {Gaussian} processes},
	year      = {2005},
	number    = {470},
	pages     = {653--668},
	volume    = {100},
	publisher = {Taylor \& Francis},
}

@article{Konomi2014,
	author = {Bledar A. Konomi and Huiyan Sang and Bani K. Mallick},
	title = {Adaptive {Bayesian} Nonstationary Modeling for Large Spatial Datasets Using Covariance Approximations},
	journal = {Journal of Computational and Graphical Statistics},
	volume = {23},
	number = {3},
	pages = {802-829},
	year  = {2014},
	publisher = {Taylor & Francis},
	doi = {10.1080/10618600.2013.812872}
}

@Article{krause2008near,
	author  = {Krause, Andreas and Singh, Ajit and Guestrin, Carlos},
	journal = {Journal of Machine Learning Research},
	title   = {Near-optimal sensor placements in {Gaussian} processes: Theory, efficient algorithms and empirical studies},
	year    = {2008},
	number  = {Feb},
	pages   = {235--284},
	volume  = {9},
}

@Article{lam2008sequential,
	author  = {Lam, Chen Quin and Notz, W.I.},
	journal = {Statistics and Applications},
	title   = {Sequential adaptive designs in computer experiments for response surface model fit},
	year    = {2008},
	number  = {1},
	pages   = {207-233},
	volume  = {6},
}

@Article{luo2021bayesian,
	author  = {Luo, Z and Sang, Huiyan and Mallick, Bani},
	journal = {Journal of Machine Learning Research},
	title   = {A {Bayesian} contiguous partitioning method for learning clustered latent variables},
	year    = {2021},
	pages   = {1-52},
	volume  = {22},
}

@book{hastie2009elements,
	title={The elements of statistical learning: data mining, inference, and prediction},
	author={Hastie, T and Tibshirani, R and Friedman, JH},
	year={2009},
	publisher={Springer},
	address={New York, NY}
}

@article{binois2018replication,
	author = {Binois, M and Huang, J and Gramacy, RB and Ludkovski, M},
	title = {Replication or exploration? {S}equential design for stochastic simulation experiments},
	journal = {Technometrics},
	volume = {27},
	number = {4},
	pages = {808--821},
	year  = {2019},
	publisher = {Taylor & Francis},
	doi = {10.1080/00401706.2018.1469433}
}

@Article{malloy2014near,
	author    = {Malloy, Matthew L and Nowak, Robert D},
	journal   = {IEEE Transactions on Information Theory},
	title     = {Near-optimal adaptive compressed sensing},
	year      = {2014},
	number    = {7},
	pages     = {4001--4012},
	volume    = {60},
	publisher = {IEEE},
}

@Article{mckay2000comparison,
	author    = {McKay, Michael D and Beckman, Richard J and Conover, William J},
	journal   = {Technometrics},
	title     = {A comparison of three methods for selecting values of input variables in the analysis of output from a computer code},
	year      = {2000},
	number    = {1},
	pages     = {55--61},
	volume    = {42},
	publisher = {Taylor \& Francis},
}

@Book{mitchell1997edition,
	author    = {Mitchell, Tom},
	title     = {Machine Learning},
	year      = {1997},
	address   = {New York, NY},
	publisher = {McGraw-Hill, Inc},
}

@Article{mu2017sequential,
	author    = {Rongji Mu and Lixiang Dai and Jin Xu},
	journal   = {Communications in Statistics - Simulation and Computation},
	title     = {Sequential design for response surface model fit in computer experiments using derivative information},
	year      = {2017},
	number    = {2},
	pages     = {1148-1155},
	volume    = {46},
	publisher = {Taylor & Francis},
}

@article{Nikolaev2016,
	author = {Nikolaev, Pavel and Hooper, Daylond and Webber, Frederick and Rao, Rahul and Decker, Kevin and Krein, Michael and Poleski, Jason and Barto, Rick and Maruyama, Benji},
	year = {2016},
	pages = {16031},
	title = {Autonomy in materials research: A case study in carbon nanotube growth},
	volume = {2},
	journal = {npj Computational Materials},
}

@Article{paciorek2006spatial,
	author    = {Paciorek, CJ and Schervish, MJ},
	journal   = {Environmetrics},
	title     = {Spatial modelling using a new class of nonstationary covariance functions},
	year      = {2006},
	number    = {5},
	pages     = {483--506},
	volume    = {17},
	publisher = {Wiley Online Library},
}

@Article{park2022,
	author  = {Chiwoo Park},
	journal = {Journal of Machine Learning Research},
	title   = {Jump {Gaussian} Process Model for Estimating Piecewise Continuous Regression Functions},
	year    = {2022},
	number  = {278},
	pages   = {1-37},
	volume  = {23},
}

@Article{pope2021gaussian,
	author    = {Pope, Christopher A and Gosling, John Paul and Barber, Stuart and Johnson, Jill S and Yamaguchi, Takanobu and Feingold, Graham and Blackwell, Paul G},
	journal   = {Technometrics},
	title     = {Gaussian process modeling of heterogeneity and discontinuities using {Voronoi} tessellations},
	year      = {2021},
	number    = {1},
	pages     = {53--63},
	volume    = {63},
	publisher = {Taylor \& Francis},
}

@InProceedings{salimbeni2017doubly,
	author    = {Salimbeni, Hugh and Deisenroth, Marc},
	booktitle = {Advances in Neural Information Processing Systems},
	title     = {Doubly stochastic variational inference for deep {Gaussian} processes},
	year      = {2017},
	editor    = {I. Guyon and U. Von Luxburg and S. Bengio and H. Wallach and R. Fergus and S. Vishwanathan and R. Garnett},
	volume    = {30},
	address   = {Long Beach, CA, USA}
}

@Article{sampson1992nonparametric,
	author    = {Sampson, PD and Guttorp, P},
	journal   = {Journal of the American Statistical Association},
	title     = {Nonparametric estimation of nonstationary spatial covariance structure},
	year      = {1992},
	number    = {417},
	pages     = {108--119},
	volume    = {87},
	publisher = {Taylor \& Francis},
}

@Article{sauer2021active,
	author  = {Sauer, Annie and Gramacy, Robert B and Higdon, David},
	journal = {Technometrics},
	title   = {Active Learning for Deep {Gaussian} Process Surrogates},
	volume  = {65},
	number  = {1},
	pages   = {4--18},
	year    = {2023},
}

@Article{schmidt2003bayesian,
	author    = {Schmidt, AM and O'Hagan, A},
	journal   = {Journal of the Royal Statistical Society: Series B},
	title     = {Bayesian inference for non-stationary spatial covariance structure via spatial deformations},
	year      = {2003},
	number    = {3},
	pages     = {743--758},
	volume    = {65},
	publisher = {Wiley Online Library},
}

@Article{taddy2011dynamic,
	author    = {Taddy, Matthew A and Gramacy, Robert B and Polson, Nicholas G},
	journal   = {Journal of the American Statistical Association},
	title     = {Dynamic trees for learning and design},
	year      = {2011},
	number    = {493},
	pages     = {109--123},
	volume    = {106},
	publisher = {Taylor \& Francis},
}

@Article{doi:10.1080/24725854.2021.1988770,
	author    = {Chiwoo Park and Peihua Qiu and Jennifer Carpena-Núñez and Rahul Rao and Michael Susner and Benji Maruyama},
	journal   = {IISE Transactions},
	title     = {Sequential adaptive design for jump regression estimation},
	year      = {2023},
	number    = {2},
	pages     = {111-128},
	volume    = {55},
	doi       = {10.1080/24725854.2021.1988770},
	eprint    = {https://doi.org/10.1080/24725854.2021.1988770},
	publisher = {Taylor & Francis},
}

\end{document}